\documentclass[11pt]{article}

\usepackage[final]{acl}

\usepackage{times}
\usepackage{latexsym}

\usepackage[T1]{fontenc}
\usepackage[utf8]{inputenc}

\usepackage{microtype}

\usepackage{inconsolata}

\usepackage{graphicx}
\usepackage{subfig}
\usepackage{booktabs}
\usepackage{amsmath,amssymb}
\usepackage{adjustbox}
\usepackage{multirow}
\usepackage{subcaption}
\usepackage{newunicodechar}
\newunicodechar{Δ}{$\Delta$}
\usepackage{algorithm}
\usepackage{algorithmic}
\usepackage{listings}
\usepackage{seqsplit}
\usepackage{xurl}
\usepackage{enumitem}
\setlist{nosep}
\title{Capturing Classic Authorial Style in Long-Form Story Generation with GRPO Fine-Tuning}

\author{
  Jinlong Liu \quad Mohammed Bahja \quad Venelin Kovatchev \quad Mark Lee \\
  School of Computer Science, University of Birmingham, United Kingdom \\
  \texttt{jxl2069@student.bham.ac.uk;\{m.bahja,v.o.kovatchev,m.g.lee\}@bham.ac.uk}
}

\begin{document}
\maketitle
\begin{abstract}
Evaluating and optimising \emph{authorial style} in long-form story generation remains challenging because style judgement is often based on human voting, which is subjective and offers no stable automatic evaluation method. We propose a two-stage pipeline. Firstly, we train a dedicated \emph{style-similarity judge} by fine-tuning a sentence-transformer with authorship-verification supervision, and calibrate its similarity scores into a bounded $[0,1]$ reward. Secondly, we use this judge as the primary reward in Group Relative Policy Optimization (GRPO) to fine-tune an \texttt{8B} story generator for style-conditioned writing, avoiding the accept/reject supervision required by Direct Preference Optimization (DPO). Across four target authors (\texttt{Mark Twain}, \texttt{Jane Austen}, \texttt{Charles Dickens}, \texttt{Thomas Hardy}), the GRPO-trained \texttt{8B} model achieves higher style scores than open-weight baselines, with an average style score of 0.893 across authors. These results suggest that authorship-verification (AV)-calibrated reward modelling provides a practical mechanism for controllable style transfer in long-form generation under moderate model size and training budget.\footnote{Code available at \url{https://github.com/Vince-Liuss/literary_style_model}}.
\end{abstract}

\section{Introduction}
Recent progress in long-form generation has made open-ended story writing a realistic use case for large language models. New writing benchmarks---including WritingBench~\citep{wu2025writingbenchcomprehensivebenchmarkgenerative}, LitBench~\citep{fein2025litbenchbenchmarkdatasetreliable}, and EQ-Bench~\citep{paech2023eqbench}---evaluate narrative quality, creativity, and coherence more systematically. However, these benchmarks primarily target \emph{general} writing ability and do not treat \emph{authorial style} as a controllable objective for long-form generation.

At the same time, ``style imitation'' is often demonstrated without rigorous evaluation. Many examples rely on shallow cues (e.g., character names, settings, or era-specific keywords) rather than higher-level stylistic properties such as narrative voice and discourse habits. This gap matters for an agentic ghostwriter setting, where a model must maintain a target voice over long outputs while still following the prompt and producing a complete story.

We address this by framing style control as an optimisation problem with an explicit scalar signal. We adopt an AV perspective~\citep{https://doi.org/10.1002/asi.21001}: instead of predicting the author, we train a \emph{style-similarity judge} that assigns a continuous similarity score to a pair of texts. This separates style evaluation from general writing quality and provides a reward that can be optimised directly.

Our method has two stages. First, we fine-tune a sentence-transformer with AV supervision and calibrate its similarity outputs into a bounded reward in $[0,1]$. Second, we use this style reward as the primary signal in Group Relative Policy Optimisation (GRPO)~\citep{shao2024deepseekmath} to fine-tune an \texttt{8B} story generator for style-conditioned long-form writing. To stabilise long generations, we add lightweight auxiliary rewards for content quality and completeness, while keeping the style judge as the main driver.

 We study four classic authors whose Library of Congress subject anchors do not overlap: \texttt{Mark Twain} (\emph{Adventure stories}), \texttt{Jane Austen} (\emph{Young women -- Fiction}), \texttt{Charles Dickens} (\emph{Historical fiction}), and \texttt{Thomas~Hardy} (\emph{Man-woman relationships -- Fiction}). Under a fixed training budget, we compare prompt-based baselines (zero-shot and few-shot) with reward-optimised models using the same reward pipeline, enabling an author-wise analysis of style alignment; our results show that mixed-language writing styles remain imperfectly captured, although the model still achieves practically useful alignment.
\begin{itemize}
    \item We propose an AV-calibrated \emph{style-similarity judge} for long-form text by fine-tuning a sentence-transformer and mapping similarity scores to a bounded $[0,1]$ reward.
    \item We use this judge as the primary reward in GRPO to fine-tune an \texttt{8B} story generator for author-conditioned writing, and report stability-sensitive settings for long-form optimisation.
    \item We benchmark prompt-based baselines and reward-optimised models under the same reward pipeline, enabling direct author-wise comparison of style alignment.
\end{itemize}

\section{Related Works}
Prior research on AV spans a range of strategies, from stylometric features to transformer-based models and prompt-driven approaches. \citet{https://doi.org/10.1002/asi.21001} offer a foundational survey of automated authorship attribution methods, outlining five key stylometric categories: (i) lexical features (e.g., word length, vocabulary richness), (ii) character-level features (e.g., character n-grams), (iii) syntactic structures (e.g., POS tag distributions, parse rules), (iv) semantic indicators (e.g., synonym usage, semantic roles), and (v) domain-specific signals (e.g., structural patterns in emails or genre-linked keywords).

Building on traditional techniques, \citet{bevendorff-etal-2019-generalizing} extend the "unmasking" method—a previously book-length-dependent AV algorithm—into a generalized form effective on shorter texts, such as articles or essays. Their adaptation preserves high precision, addressing the impracticality of long-text dependency in real-world scenarios.

Recent efforts focus on disentangling style from content. \citet{wang-etal-2023-authorship} show that even when semantic content is obfuscated through masking and paraphrasing, style-based features remain discriminative. In parallel, \citet{nguyen-etal-2023-improving} evaluate transformer architectures on long fanfiction narratives, employing balanced sampling and sliding-window encoding. Their BigBird pipeline achieves a 4.8\% equal error rate and a PAN21 score of 0.9596, outperforming Longformer, RoBERTa, and ELECTRA.

Prompt-based AV has recently gained attention. \citet{hung-etal-2023-wrote} introduce PromptAV, leveraging step-by-step reasoning with stylometric cues. Their method surpasses standard prompting strategies (e.g., Chain-of-Thought, PS+), achieves 66.7\% accuracy in few-shot IMDb62 classification. Meanwhile, \citet{sawatphol-etal-2024-addressing} emphasize evaluation robustness through Heterogeneity-Informed Topic Sampling (HITS), which increases topic diversity and yields stable rankings across random seeds, with a Kendall rank correlation of 0.92.

Interpretability has also become an important focus in AV. \citet{ramnath-etal-2025-cave} introduce CAVE, an explanation framework that produces linguistic rationales based on surface-level features such as punctuation, syntactic structure, and stylistic variation. Evaluated across three AV benchmarks (IMDB62, BLOG-AUTH, and FANFICTION), CAVE achieves competitive accuracy while significantly improving human interpretability. In contrast, \citet{michel-etal-2024-distinguishing} investigate AV in the context of literary dialogue, highlighting the difficulty of distinguishing character voice from true authorial style in fictional texts.

For agentic model training, recent work on GRPO has shown encouraging results. This method involves loading the model twice—once as a reference model to measure divergence, and once as the trainable model—to iteratively update the policy based on a custom reward function. GRPO has been successfully applied to models like \texttt{DeepSeek-R1} without requiring a separate critic network. By estimating reward baselines from grouped sample scores, GRPO significantly reduces the computational overhead compared to traditional methods such as Proximal Policy Optimization (PPO) \citep{shao2024deepseekmath}. 

Unlike earlier approaches that use DPO \citep{rafailov2023direct}, which depends on explicitly labelled accept/reject pairs, GRPO allows for greater output diversity. This flexibility is especially critical for open-ended creative tasks like story generation, where multiple stylistic and narrative variations may be acceptable or even desirable.

This work builds on these developments with a focus on writing-style AV. We investigate whether large language models, when fine-tuned with targeted reward signals, can consistently learn and reproduce the stylistic signatures of classic authors in long-form fiction by applying new agentic training method instead of a lot of sample text.

\section{Methodology}
In this section, we introduce a method to identify writing style similarity for direct use as a reward function in agentic ghostwriter training. Traditional stylometry methods, such as Burrows’ Delta~\citep{10.1093/llc/17.3.267} and the TED Algorithm~\citep{zhang1989simple}, rely on token-level statistics and work well in controlled or short-text settings. However, for long-form generation these signals become less reliable: style is expressed through dispersed cues (voice, rhythm, discourse patterns) and token-level distances are easily confounded by topic and content.
To build a style reward model that is usable in agentic fine-tuning, we require:
\begin{enumerate}
    \item \textbf{Reward smoothness under partial imitation:} The score should increase gradually as style improves, rather than collapsing into near-binary decisions. This is necessary to provide usable gradients during early training when the model only approximates the target voice.
    \item \textbf{Robustness across lengths and prompts:} The reward must behave consistently across different story lengths and diverse prompts, including unseen generations produced during optimisation.
    \item \textbf{Scalable similarity supervision:} Training requires large numbers of text pairs with interpretable similarity labels, produced at low cost and without leaking topic cues that allow superficial shortcut solutions.
\end{enumerate}

The remainder of this section describes how we meet these requirements using controlled augmentation to generate graded similarity labels, and how we validate and fine-tune pretrained sentence encoders into a calibrated style-similarity judge.

\subsection{Baseline Analysis of Style Similarity Models}
As an initial step, we explore a sentence-transformer baseline that uses pretrained embeddings to compute pairwise similarity between text chunks. Each pair is assigned a binary label indicating whether the two chunks are written by the same author. We then analyse the resulting similarity score distributions to assess whether off-the-shelf embeddings provide sufficient separation, or whether style-specific fine-tuning is necessary. To validate the reliability of sentence-transformer embeddings for style similarity detection, we conduct a controlled experiment using authors with two different books within the same subject across four subjects. Texts are segmented into 500–3000-word chunks, and we sample 3{,}000 pairs per subject with balanced composition as a stability test. We construct both intra-author and cross-author pairs to isolate stylistic consistency while reducing content-level confounds, and interpret the resulting similarity scores as a proxy for authorial style similarity.

We evaluated four representative sentence-transformer models for their architectural diversity and optimization for sentence-level semantic tasks and they both have context length in 8192. Four models—\texttt{BGE-M3}\footnote{\url{https://huggingface.co/BAAI/bge-m3}}, \texttt{\seqsplit{ModernBERT-embed-base-legal-MRL}}\footnote{\url{https://huggingface.co/AdamLucek/ModernBERT-embed-base-legal-MRL}}, \texttt{\seqsplit{Nomic-embed-text-v1.5}}\footnote{\url{https://huggingface.co/nomic-ai/nomic-embed-text-v1.5}}, and \texttt{\seqsplit{GTE-large-en-v1.5}}\footnote{\url{https://huggingface.co/Alibaba-NLP/gte-large-en-v1.5}}. They all been fine-tuned on semantic research content.

Each model was assessed across multiple chunk sizes using several statistical criteria (mean Δ, IQR overlap, standard deviation), and performance was benchmarked before and after contrastive fine-tuning. Results show that while pre-trained models exhibit varying degrees of sensitivity to authorial style, their baseline performance is limited—highlighting the need for task-specific adaptation. Fine-tuning consistently improved both discriminative power and distributional separation, confirming the viability of this approach for style-oriented representation learning.

\begin{table*}[t]
\centering
\resizebox{0.9\textwidth}{!}{%
\begin{tabular}{@{}lcccc@{}}
\toprule
\textbf{Model} &
\textbf{Cross Author $\pm$ SD} &
\textbf{Same Author $\pm$ SD} &
\textbf{$\Delta$} &
\textbf{IQR Overlap} \\ \midrule
\texttt{BGE-M3} & 0.624 $\pm$ 0.046 & 0.675 $\pm$ 0.057 & +0.051 & 25.8\% \\
\texttt{GTE-large-en-v1.5} & 0.664 $\pm$ 0.049 & 0.720 $\pm$ 0.068 & +0.057 & 34.5\% \\
\texttt{ModernBERT-embed-base-legal-MRL} & 0.616 $\pm$ 0.063 & 0.686 $\pm$ 0.067 & +0.070 & 18.6\% \\
\texttt{Nomic-embed-text-v1.5} & 0.736 $\pm$ 0.053 & 0.787 $\pm$ 0.052 & +0.051 & 18.7\% \\
\bottomrule
\end{tabular}}
\caption{Baseline embedding performance averaged across chunk sizes (500--3000).
Values report mean similarity $\pm$ standard deviation for cross-author vs.\ same-author pairs; $\Delta = \text{Same} - \text{Cross}$.
\textbf{Higher is better for $\Delta$}, while \textbf{lower is better for IQR Overlap}.
IQR Overlap is the percentage intersection of the two 25--75\,\% percentile ranges (relative to the smaller IQR).
Per chunk size: $n_{\text{cross}}=n_{\text{same}}=1500$.}
\label{tab:baselines}
\end{table*}

\begin{table*}[t]
\centering
\resizebox{0.9\textwidth}{!}{%
\begin{tabular}{@{}lcccc@{}}
\toprule
\textbf{Model} &
\textbf{Cross Author $\pm$ SD} &
\textbf{Same Author $\pm$ SD} &
\textbf{$\Delta$} &
\textbf{IQR Overlap} \\ \midrule
\texttt{BGE-M3} & 0.371 $\pm$ 0.190 & 0.656 $\pm$ 0.154 & +0.285 & 0.1\% \\
\textbf{\texttt{GTE-large-en-v1.5}} & \textbf{0.273 $\pm$ 0.239} & \textbf{0.679 $\pm$ 0.178} & \textbf{+0.406} & \textbf{0.0\%} \\
\texttt{ModernBERT-embed-base-legal-MR} & 0.330 $\pm$ 0.194 & 0.636 $\pm$ 0.136 & +0.306 & 0.0\% \\
\texttt{Nomic-Embed-v1.5} & 0.431 $\pm$ 0.175 & 0.677 $\pm$ 0.131 & +0.246 & 4.0\% \\
\bottomrule
\end{tabular}}
\caption{Fine-tuned embedding performance averaged across chunk sizes (500--3000).
Fine-tuning increases the same-vs-cross separation ($\Delta$) and reduces IQR overlap.
\textbf{Higher is better for $\Delta$}, while \textbf{lower is better for IQR Overlap}.
$\Delta = \text{Same} - \text{Cross}$; IQR Overlap is defined as in Table~\ref{tab:baselines}.
Per chunk size: $n_{\text{cross}}=n_{\text{same}}=1500$.}
\label{tab:fine_tuned}
\end{table*}

Table~\ref{tab:baselines} reports baseline results averaged across chunk sizes. All models show limited separation between same-author and cross-author pairs (small $\Delta$ and non-trivial IQR overlap), indicating substantial distributional overlap. \texttt{\seqsplit{ModernBERT-embed-base-legal-MRL}} yields the largest margin ($\Delta=+0.070$) and lowest overlap (18.6\%), while \texttt{\seqsplit{GTE-large-en-v1.5}} and \texttt{BGE-M3} show smaller margins ($\Delta=+0.057$ and $+0.051$) with higher overlap (34.5\% and 25.8\%). \texttt{\seqsplit{Nomic-embed-text-v1.5}} achieves the highest absolute similarities for both pair types, but still only a modest gap ($\Delta=+0.051$) with 18.7\% overlap. Full per-chunk-size results are provided in Appendix~\ref{Sec:appendix a}.

Overall, off-the-shelf embedding models provide weak-to-moderate author discrimination in our setting, motivating task-specific fine-tuning to amplify stylistic signals beyond general semantic similarity and to reduce overlap between score distributions.
\subsection{Constructing a Style-Controlled Dataset for Reward Modelling}
\label{sec:reward_model}
\begin{figure}[t]
    \centering
    \includegraphics[width=0.5\textwidth]{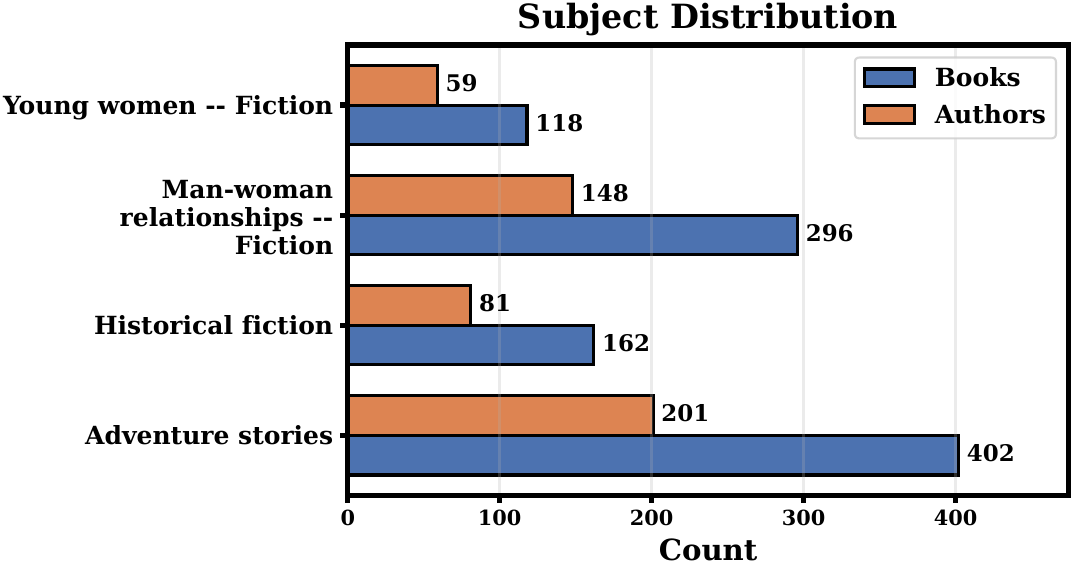}
    \caption{Subjects \& Authors Distributions}
    \label{fig:subjects}
\end{figure}

We construct a pairwise dataset that provides graded supervision for learning a continuous style-similarity function while controlling for topic. Our pipeline has two steps: (i) generate controlled ``refills'' of real literary chunks by masking and regenerating sentences, and (ii) form labelled text pairs within each subject using author-disjoint splits.
\paragraph{Chunk Refilling via Sentence Masking.}
Given an original chunk $C$ consisting of at least ten sentences, we sample a masking ratio $r \in \{0.1,0.2,\dots,0.9\}$ and mask a subset of sentences (by length-based selection). We then regenerate the masked sentences sequentially with \texttt{\seqsplit{GPT-oss-20B}} \citep{openai2025gptoss120bgptoss20bmodel} to obtain a refilled chunk $C'(r)$. For original--refilled pairs $(C, C'(r))$, a simple heuristic is that content overlap decreases with $r$ (approximately $1-r$), while higher-level stylistic traits may remain partially preserved.
\paragraph{Source Corpus and Subject Control.}
We draw texts from Project Gutenberg and restrict the corpus to four metadata subjects to reduce topical confounds: \textit{Adventure stories}, \textit{Historical fiction}, \textit{Young women -- Fiction}, and \textit{Man-woman relationships -- Fiction}. The resulting corpus contains 489 authors and 978 books; Figure~\ref{fig:subjects} summarises the subject distribution. We apply the refilling procedure to each chunk to create multiple controlled variants per title and author.

\begin{figure*}[t]
    \centering
    \includegraphics[width=0.9\linewidth]{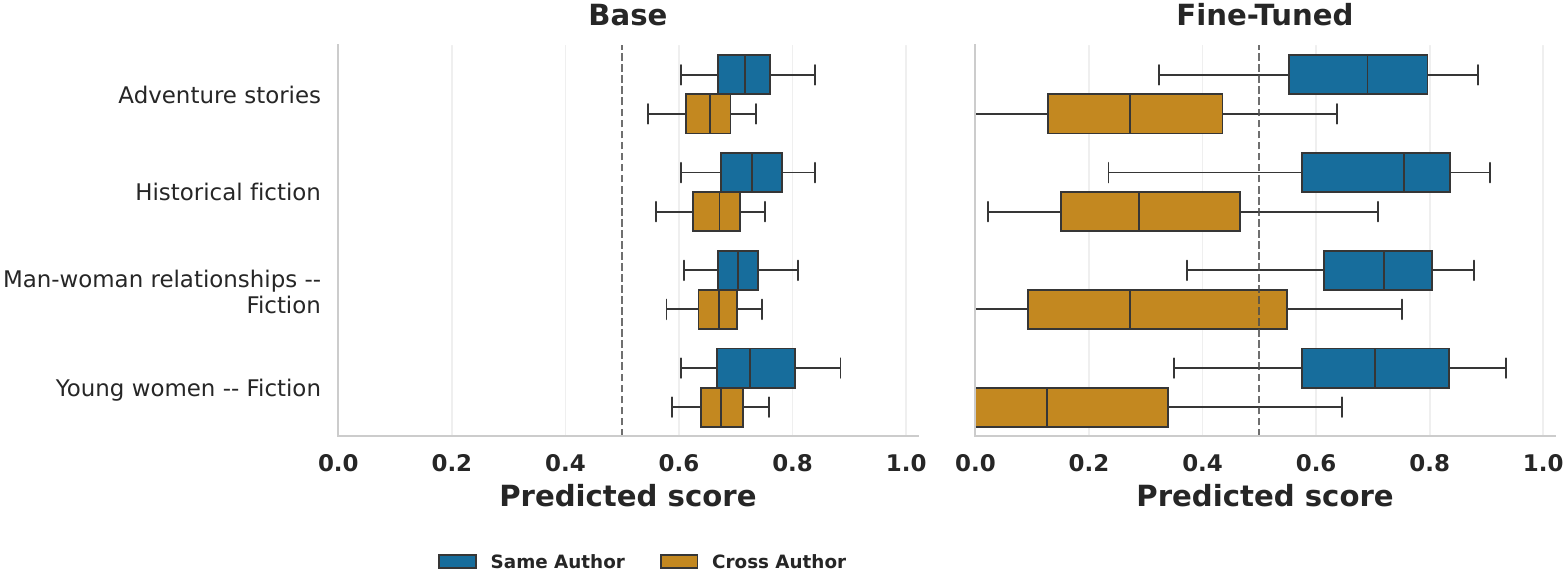}
    \caption{Subjects performance of \texttt{GTE-large-en-v1.5}}
    \label{fig:subjects_GTE}
\end{figure*}

\paragraph{Pairwise Dataset Construction.}
For each subject, we partition authors into disjoint train/validation/test splits and construct pairs only within the same subject. This yields author-disjoint validation and test sets, so evaluation reflects zero-shot generalisation to unseen authors.

We reserve extreme labels for high-confidence supervision using only original chunks. Positive pairs with $s{=}1.0$ are formed by matching original chunks from different titles by the same author within a subject, and negative pairs with $s{=}0.0$ are formed by matching original chunks from different authors within the subject. We do not use original chunks to populate intermediate labels: the heuristic similarity $1-r$ is only directly defined for original--refilled pairs, and using originals throughout would overuse a small set of chunks, reducing pair diversity and increasing memorisation risk.

Intermediate labels are therefore constructed from refilled--refilled pairs under the same-subject, different-title constraint. We omit the midpoint label $s{=}0.5$ because it provides an ambiguous target that can encourage lexical matching rather than sentence-level style cues. For $s{>}0.5$ we use same-author refilled pairs, and for $s{<}0.5$ we use cross-author refilled pairs. Since all refills are produced by the same generator, even cross-author refills share some lexical regularities, making mid-range supervision more realistic than a strict 0/1 regime.

Given masking ratios $(r_1,r_2)$ for a refilled--refilled pair, we assign a continuous target $x$ and discretise it to the nearest tenth:
\begin{eqnarray}
x &=& \Big((1-r_1^{2})(1-r_2^{2})\Big)^{2}, \label{eq:pair_score_x}\\
s &=& \frac{1}{10}\left\lfloor 10x + \frac{1}{2}\right\rfloor. \label{eq:pair_score_s}
\end{eqnarray}
This construction supports calibration analyses of the learned judge, including whether the model’s implicit midpoint deviates from the ideal boundary at $0.5$.

Overall, the procedure yields 100K training pairs and approximately 10K pairs each for validation and test, with balanced coverage over subjects and score bins.\footnote{Dataset: \url{https://huggingface.co/datasets/VibrantVista/style-judge-dataset}}.

\paragraph{Performance Gains and Model Selection for Agentic Training.}
Table~\ref{tab:fine_tuned} summarises fine-tuned performance on the held-out benchmark (full per-chunk results and boxplots in Appendix~A). Fine-tuning increases the same--vs--cross margin for every model and sharply reduces distributional overlap. \texttt{\seqsplit{GTE-large-en-v1.5}} achieves the largest average separation ($\Delta{=}+0.406$) with 0.0\% IQR overlap, while \texttt{\seqsplit{ModernBERT-embed-base-legal-MRL}} and \texttt{BGE-M3} show similarly strong gains ($\Delta{=}+0.306$ and $+0.285$) with near-zero overlap. \texttt{\seqsplit{Nomic-embed-text-v1.5}} improves more moderately ($\Delta{=}+0.246$) and retains a small residual overlap (4.0\%).

\begin{figure}[ht]
    \centering
    \includegraphics[width=0.85\linewidth]{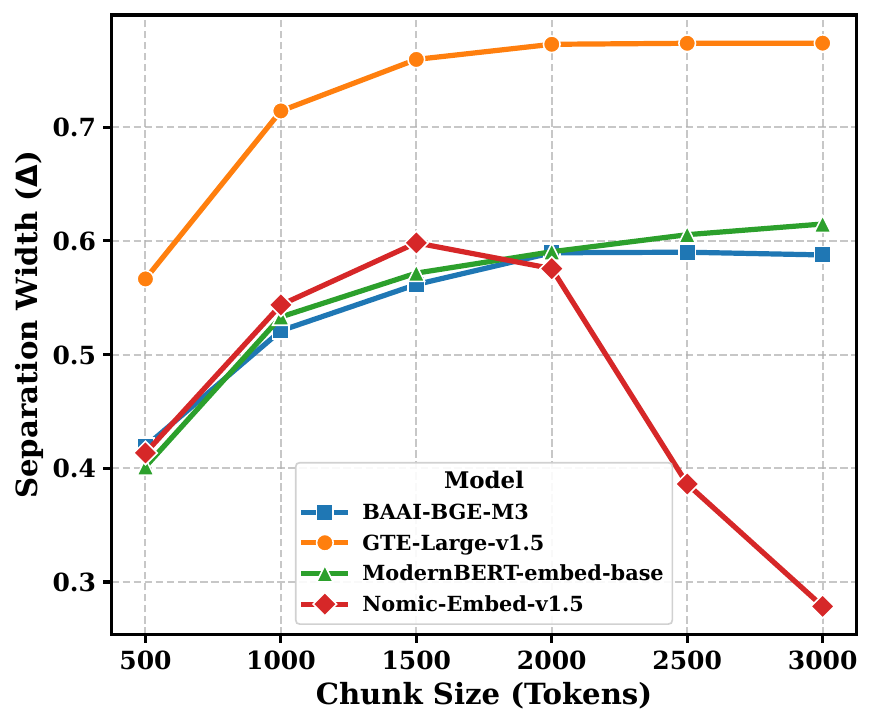}
    \caption{Separation width vs.\ chunk size. Separation margin $\Delta$ (Same$-$Cross) as a function of chunk length. \texttt{\protect\seqsplit{GTE-large-en-v1.5}} consistently achieves the largest margin, peaking around $\Delta\approx0.77$ for chunk sizes $\ge$2000 tokens. \texttt{BGE-M3} and \texttt{\protect\seqsplit{ModernBERT-embed-base-legal-MRL}} improve steadily with chunk size, whereas \texttt{\protect\seqsplit{Nomic-embed-text-v1.5}} degrades beyond $\sim$2000 tokens, reducing same--vs--cross separation.}
    \label{fig:separation}
\end{figure}

\begin{figure}[ht]
    \centering
    \includegraphics[width=0.85\linewidth]{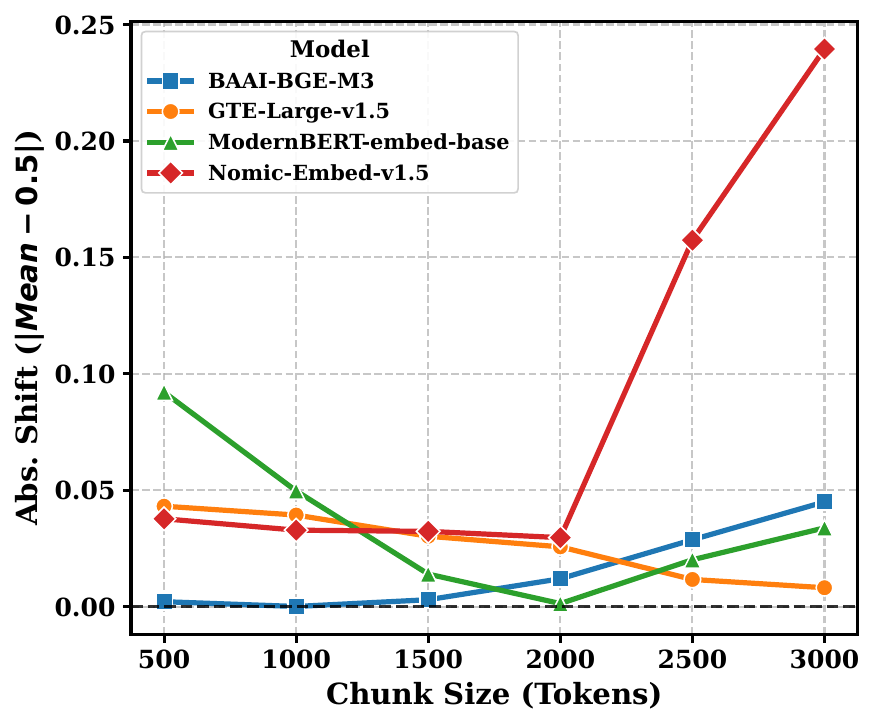}
    \caption{Midpoint shift by chunk size. Absolute deviation of the learned midpoint from the ideal decision boundary (0.5); lower is better. \texttt{BGE-M3} and \texttt{\protect\seqsplit{GTE-large-en-v1.5}} remain well-calibrated (typically $|\text{shift}|<0.05$) across most chunk sizes. \texttt{\protect\seqsplit{ModernBERT-embed-base-legal-MRL}} shows mild drift at shorter chunks, while \texttt{\protect\seqsplit{Nomic-embed-text-v1.5}} diverges as context length increases.}
    \label{fig:midpoint}
\end{figure}

To select the reward signal for agentic training, we compare models along three criteria: (1) alignment with the intended decision boundary at $0.5$, (2) effective separation range (Q75$_{\text{same}}$ -- Q25$_{\text{cross}}$), and (3) robustness to cross-author outliers near the boundary. Figure~\ref{fig:separation} and \ref{fig:midpoint} shows that fine-tuning re-centers all models, reducing the positive bias observed in the pretrained baselines (shifts $>+0.15$). \texttt{BGE-M3} is the most precisely centred (shift $+0.015$), but \texttt{\seqsplit{GTE-large-en-v1.5}} provides the strongest overall reward profile: it remains close to the boundary (shift $-0.026$) while offering a substantially wider usable signal range (0.73 vs.\ 0.47--0.55 for the other models). It also exhibits the highest boundary robustness, suppressing false-positive cross-author scores that can otherwise enable reward hacking.
Finally, Figure~\ref{fig:subjects_GTE} shows that \texttt{\seqsplit{GTE-large-en-v1.5}} maintains clear separation across all subjects after fine-tuning, supporting stable reward signals under subject-wise evaluation. Subject-wise plots for the remaining models are provided in Appendix~\ref{Sec:appendix a} Figure~\ref{fig:subjects_others}.

\section{Dataset Preparation}
This section describes the data construction pipeline for training a story-writing agent with GRPO. We build two datasets: (i) a mixed-task SFT set that teaches general story completion and style imitation, and (ii) a GRPO set that applies reward-driven optimisation for style control. We focus on four target books as style references:
\begin{itemize}
    \item Mark Twain, \textit{Adventures of Huckleberry Finn}
    \item Charles Dickens, \textit{A Tale of Two Cities}
    \item Jane Austen, \textit{Pride and Prejudice}
    \item Thomas Hardy, \textit{Tess of the d'Urbervilles: A Pure Woman}
\end{itemize}

\paragraph{SFT dataset.}
We build the SFT starting point by adapting the 300K prompt–story pairs of \citet{fan-etal-2018-hierarchical}. Because the original prompts vary widely in tone and length, we curate 50 plot prompts with fixed constraints (1200–1500 words; resolved ending; final line \texttt{THE END}) to teach reliable long-form completion. Using these prompts, we sample 80K $\sim$1{,}500-word stories from \texttt{GPT-oss-120B} \citep{openai2025gptoss120bgptoss20bmodel}, then produce style-imitation variants for each target book via few-shot prompting that conditions on reference chunks. We filter all generations with the selected style judge (Section~\ref{sec:reward_model}) and fine-tune the base model on the resulting mixture to obtain a strong general \footnote{Dataset:\url{https://huggingface.co/datasets/VibrantVista/story-style-SFT-dataset}}
. Prompt templates and the scoring rubric are provided in Appendix~\ref{Sec:Appendix b}.
\paragraph{GRPO dataset.}
For GRPO, we construct an author-specific dataset for each target book. Following evidence that single-work conditioning improves style fidelity \citep{wilmot2022great}, we treat each target book as the style corpus and segment it into 1{,}500-word chunks to match the length constraint. Each instance pairs one of the 50 plots with a sampled reference chunk and asks the model to realise the plot in the target style under the same constraints as SFT, augmented with explicit \emph{Author/Title} fields. We hold out 10 plots per author for evaluation (disjoint from training plots), yielding 1,800 training prompts and 200 test prompts per author. Reusing the SFT plots keeps the narrative content familiar, reducing instability from unseen plot structures during GRPO\footnote{Dataset:\url{https://huggingface.co/datasets/VibrantVista/grpo-style-training}}
. Prompt formats are listed in Appendix~\ref{Sec:Appendix b}.

\begin{figure*}[t]
  \centering
  \subfloat[Training Reward Progress by $\beta$ Value\label{fig:reward}]{
    \includegraphics[width=0.45\linewidth]{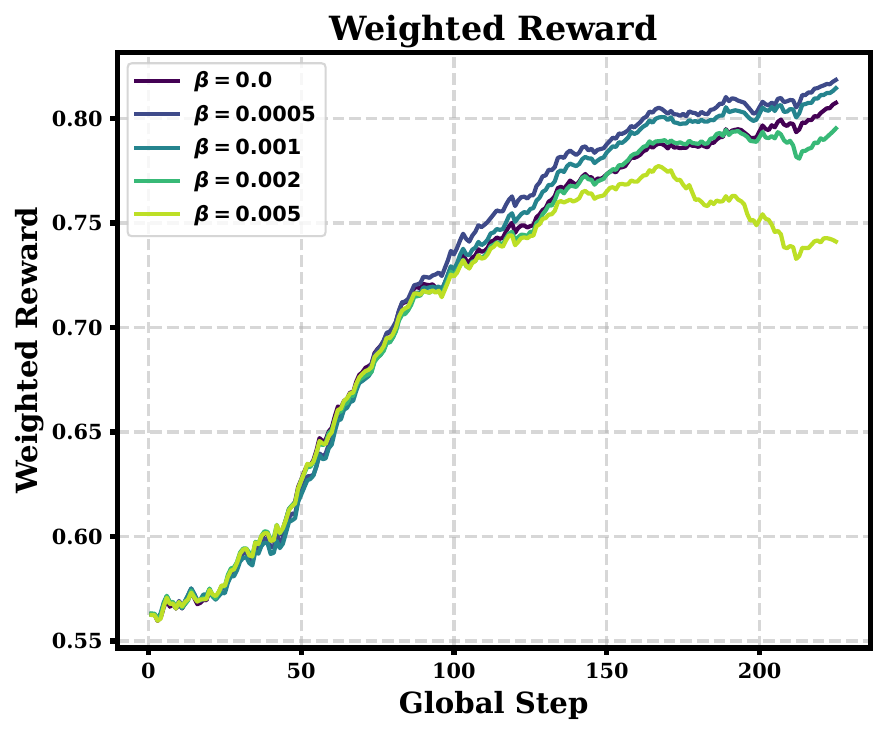}
  }
  \hfill
  \subfloat[Training KL Value by Beta Value\label{fig:kl}]{
    \includegraphics[width=0.45\linewidth]{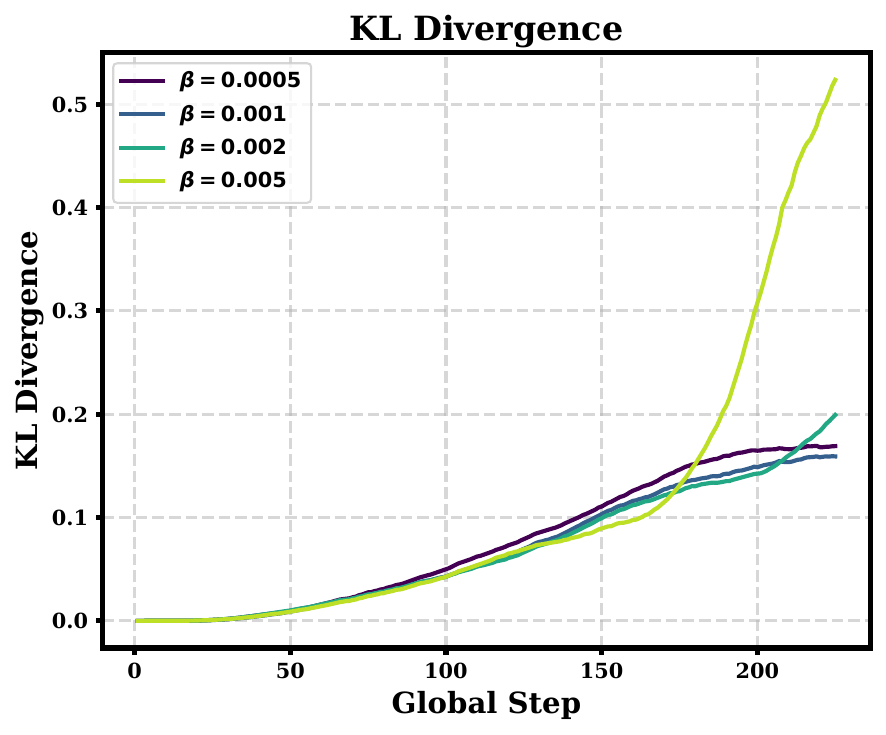}
  }
  \caption{Reward and KL-divergence trajectories during training. \textbf{(a)} Smaller $\beta$ generally yields higher final rewards. \textbf{(b)} With smaller $\beta$, KL increases smoothly and consistently, indicating stable policy updates, whereas $\beta\in\{0.002,0.005\}$ shows late-stage instability. Overall, these trends suggest that a weak KL constraint is sufficient—and often preferable—for this task. \emph{Note:} when $\beta=0$, we do not load a reference model, so KL is not logged.}
  \label{fig:GRPO_plot}
\end{figure*}

\section{Experiment}
In this section, we test GRPO for training language models on creative writing tasks. GRPO uses group-based reward estimation, which encourages exploration during policy updates and is suitable for open-ended story generation. In this experiment, the base writer model is first obtained from the SFT stage, and then further aligned to the target author style with GRPO.

\paragraph{Reward Function Setup.}
We use three reward components: (i) a calibrated \textit{style} reward, (ii) a lightweight \textit{content} reward, and (iii) a \textit{completeness} reward. The final reward is a weighted sum to keep the main objective on style alignment, while maintaining narrative quality and basic generation constraints.

\textit{Style reward.}
During training, the fine-tuned sentence-transformer judge produces a raw similarity score $S_{\text{sim}}$ between each generated story and its in-batch reference, and we use this score directly as the GRPO style reward: $r_{\text{style}} = S_{\text{sim}}$. For analysis, we report benchmark calibration statistics and treat $Q75_{\text{same}}$ (the 75th percentile of same-author similarity) as a practical upper reference, i.e., an approximate ``1.0’’ style level under our setting where generated stories are compared to author reference text with different narrative content. Since the judge is initialised from an embedding model trained on semantic retrieval data, the similarity signal may still reflect residual lexical/semantic matching in addition to stylistic cues.

\textit{Content reward.}
We evaluate narrative quality, including coherence, characterisation, and readability, using \texttt{openchat-3.5-0106}~\citep{wang2023openchat}. The judge returns a single integer score in $\{1,\dots,9\}$ under a fixed rubric, which is then linearly rescaled to $[0,0.9]$. The evaluation prompt and rubric are provided in Appendix~\ref{Sec:Appendix b}.

\textit{Completeness reward.}
This reward checks whether the story meets the length requirement and ends with the required marker. Let $w$ be the word count, with $W_{\text{lo}}{=}1200$, $W_{\text{hi}}{=}1500$, $W_{\min}{=}500$, and $W_{\text{over}}{=}1600$. We assign zero reward if $w<W_{\min}$. Otherwise, we compute the length score:
\begin{equation}
\ell(w)=
\begin{cases}
w/W_{\text{lo}} & (W_{\min}\le w < W_{\text{lo}}),\\
1               & (W_{\text{lo}}\le w \le W_{\text{over}}),\\
1-\sqrt{\frac{w-W_{\text{over}}}{W_{\text{hi}}}} & (w>W_{\text{over}}),
\end{cases}
\label{eq:length_score}
\end{equation}
and enforce the required ending:
\begin{equation}
r_{\text{comp}}=\ell(w)\cdot \mathbf{1}\!\left[\text{suffix}(y)=\texttt{THE END}\right],
\label{eq:completeness_reward}
\end{equation}
where $\text{suffix}(y)$ denotes the final non-empty line of the output.

\textit{Final reward.}
We combine the three rewards as:
\begin{equation}
r=0.6\,r_{\text{style}}+0.3\,r_{\text{content}}+0.1\,r_{\text{comp}},
\end{equation}
so the model mainly optimises for stylistic alignment, while still receiving signals for narrative quality and basic completion constraints.

\begin{table}[t]
\centering
\small
\setlength{\tabcolsep}{4pt}
\begin{tabular}{@{}llc@{}}
\toprule
\textbf{Author} & \textbf{Model} & \textbf{Style Score ($\pm$ SD)} \\
\midrule

\multirow{5}{*}{\textbf{Jane Austen}}
& \textbf{FT-Model}           & \textbf{0.984 $\pm$ 0.030} \\
& \texttt{Gemma3-27B}         & 0.283 $\pm$ 0.086 \\
& \texttt{Qwen2.5-32B}        & 0.422 $\pm$ 0.110 \\
& \texttt{Llama3-70B}         & 0.356 $\pm$ 0.083 \\
& \texttt{GPT-oss-120B}       & 0.521 $\pm$ 0.110 \\
\midrule

\multirow{5}{*}{\textbf{Charles Dickens}}
& \textbf{FT-Model}           & \textbf{0.807 $\pm$ 0.148} \\
& \texttt{Gemma3-27B}         & 0.298 $\pm$ 0.099 \\
& \texttt{Qwen2.5-32B}        & 0.278 $\pm$ 0.092 \\
& \texttt{Llama3-70B}         & 0.271 $\pm$ 0.088 \\
& \texttt{GPT-oss-120B}       & 0.370 $\pm$ 0.126 \\
\midrule

\multirow{5}{*}{\textbf{Thomas Hardy}}
& \textbf{FT-Model}           & \textbf{0.819 $\pm$ 0.102} \\
& \texttt{Gemma3-27B}         & 0.337 $\pm$ 0.110 \\
& \texttt{Qwen2.5-32B}        & 0.275 $\pm$ 0.091 \\
& \texttt{Llama3-70B}         & 0.282 $\pm$ 0.088 \\
& \texttt{GPT-oss-120B}       & 0.511 $\pm$ 0.121 \\
\midrule

\multirow{5}{*}{\textbf{Mark Twain}}
& \textbf{FT-Model}           & \textbf{0.962 $\pm$ 0.061} \\
& \texttt{Gemma3-27B}         & 0.496 $\pm$ 0.133 \\
& \texttt{Qwen2.5-32B}        & 0.473 $\pm$ 0.096 \\
& \texttt{Llama3-70B}         & 0.477 $\pm$ 0.119 \\
& \texttt{GPT-oss-120B}       & 0.663 $\pm$ 0.138 \\
\bottomrule
\end{tabular}
\caption{Zero-shot author-wise style score (mean $\pm$ SD; higher is better) on test set. Values are scaled Eq~\ref{eq:style_calib}.}
\label{tab:zero_shot}
\end{table}

\paragraph{Experimental Setup:}  
All experiments were conducted on a system with 2$\times$ AMD EPYC\textsuperscript{\textregistered} 9554 CPUs (64 cores each), 768\,GB RAM, and 4$\times$ NVIDIA H100 GPUs (80\,GB VRAM each). We use the DrGRPO loss~\citep{liu2025understandingr1zeroliketrainingcritical} and set \texttt{Num\_generation} to 16 to ensure multiple samples per prompt. During initial exploration, we train for one epoch with learning rate $1\times10^{-6}$ and gradient clipping (\texttt{max\_grad\_norm}) set to 0.35. We train four separate GRPO models, one per target author, each initialised from the same SFT base model. Across all runs, the reward pipeline and optimisation settings are kept fixed except for the author-specific style reference and the corresponding GRPO fine-tuning run.
\paragraph{Tuning the Beta Parameter for Stability:}
We tune the KL coefficient beta ($\beta$) on the \texttt{Mark Twain} setting. Following DeepSeek~\citep{Guo_2025}, we start from $\beta{=}0.001$ and sweep $\beta \in \{0,\;0.0005,\;0.001,\;0.002,\;0.005\}$. We observe that larger $\beta$ values ($\beta \ge 0.002$) tend to yield limited reward improvement while KL grows rapidly, and training quality often degrades mid-run. Figures~\ref{fig:reward} and~\ref{fig:kl} show that smaller $\beta$ produces higher rewards and smoother optimisation curves in our sweep. Notably, $\beta{=}0.0005$ slightly outperforms $\beta{=}0$, suggesting that a weak KL constraint can improve training dynamics for style-conditioned long-form generation.

Based on this sweep, we set $\beta{=}5{\times}10^{-4}$ for all final runs and fine-tune one GRPO model per author. We use learning rate $1{\times}10^{-6}$ for \texttt{Mark Twain}, while the remaining authors use $1.5{\times}10^{-6}$. We evaluate our GRPO-trained models against open-weight baselines---\texttt{Gemma3-27B}~\citep{gemmateam2025gemma3technicalreport}, \texttt{Qwen2.5-32B}~\citep{qwen2.5}, and \texttt{Llama3.3-70B}~\citep{grattafiori2024llama3herdmodels}---as well as a larger reference model, \texttt{GPT-oss-120B}~\citep{openai2025gptoss120bgptoss20bmodel}. For each generated story, the style judge outputs a raw similarity $s$, which we calibrate using benchmark quartiles and clamp to $[0,1]$. Let $a=Q25_{\text{cross}}$ and $b=Q75_{\text{same}}$; we compute:
\begin{equation}
S_{\text{style}}(s)=\left[\frac{s-a}{b-a}\right]{0}^{1},
\label{eq:style_calib}
\end{equation}
where $[x]_{0}^{1} = \max(0, \min(1, x))$. Tables~\ref{tab:zero_shot} and~\ref{tab:few_shot} report author-wise means and standard deviations of $S_{\text{style}}$ across prompts under zero-shot and few-shot settings.

For our baselines, scores are most consistent on \texttt{Mark Twain}, and \texttt{GPT-oss-120B} achieves the highest overall baseline performance. For our GRPO-trained models, \texttt{Jane Austen} and \texttt{Mark Twain} yield the highest style alignment ($0.984$ and $0.962$), while \texttt{Thomas Hardy} and \texttt{Charles Dickens} remain lower ($0.819$ and $0.807$). This performance gap suggests that linguistically heterogeneous features present a higher difficulty for style alignment, making them harder for the model to capture. Specifically, the model struggles to retain distinct signals such as French insertions in \texttt{Charles Dickens} (\textit{A Tale of Two Cities})~\citep{133fcc61-677b-3440-ac08-ad515f291f17} and dialect or archaic vocabulary in \texttt{Thomas Hardy} (\textit{Tess of the d’Urbervilles})~\citep{10.1093/acprof:oso/9780198122616.001.0001}.

As an external validity check, the evaluation texts for \textit{Pride and Prejudice} (\texttt{Jane Austen}) and \textit{Tess of the d’Urbervilles} (\texttt{Thomas Hardy}) are excluded from the judge fine-tuning data, so the reported scores reflect generalisation to unseen titles rather than memorisation.
%\footnote{All models are available at \url{https://huggingface.co/collections/VibrantVista/literary-style-grpo-models}.}.

\section{Conclusion}
This work presents a pipeline for agentic long-form story generation with controllable authorial style. We train a style-similarity judge with authorship-verification supervision and calibrate its similarity outputs into a bounded $[0,1]$ reward. We then use this calibrated style reward as the primary signal in GRPO to fine-tune an \texttt{8B} story generator for style-conditioned writing. Under the same reward pipeline, the GRPO-trained models improve style alignment over open-weight baselines across all four target authors. These results suggest that a calibrated style judge can provide a practical optimisation signal for GRPO-based author-style adaptation at moderate model scale.
\section*{Limitations}
Our training pipeline is designed to be lightweight, but this also constrains supervision and data coverage. Firstly, we use a small fixed plot set (50 plots) across SFT and GRPO; this limits narrative diversity and can encourage repeated high-reward structures (e.g., similar openings). Secondl, the prompt are in same format which may lead to weak performance when changing prompt format, diversity of prompt type will help to reduce this issue.

\section*{Ethics Statement}
This work studies controllable stylistic generation in a transparent research setting. The target authors are historical and public-domain, and the model is evaluated and reported as an approximation of stylistic patterns rather than an attempt to impersonate individuals. The intended use is research and creative-writing support, with clear disclosure that outputs are machine-generated and may reflect biases or artifacts from training data and reward design.

\section*{Acknowledgements}
We are deeply grateful to the University of Birmingham for providing access to the Baskerville HPC system and the computational resources that made this research possible.

\bibliography{custom}

\appendix
\onecolumn
\section{Detailed Result}\label{Sec:appendix a}

\begin{figure}[ht]
    \centering
    \includegraphics[width=0.8\linewidth]{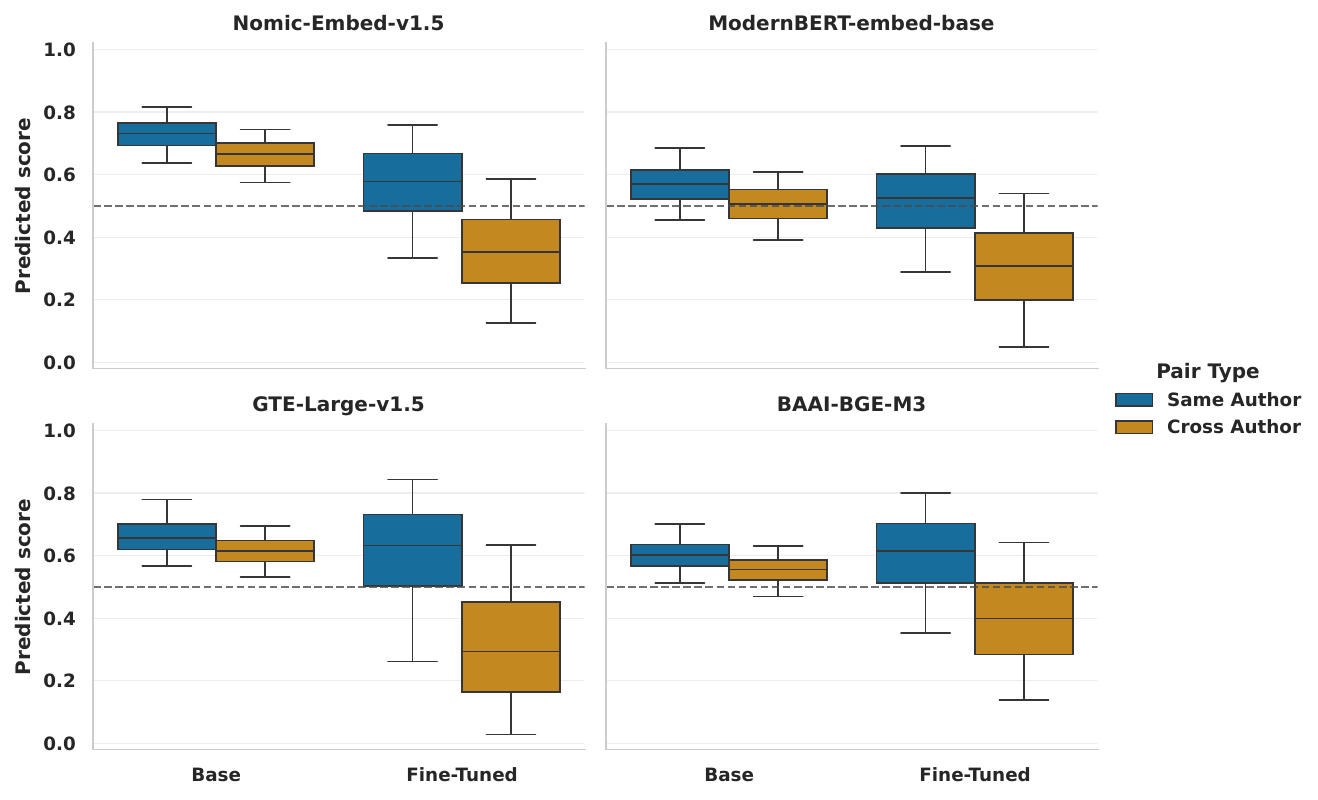}
    \caption{Result of Chunk size 500}
    \label{fig:chunk500}
\end{figure}

\begin{figure}[ht]
    \centering
    \includegraphics[width=0.8\linewidth]{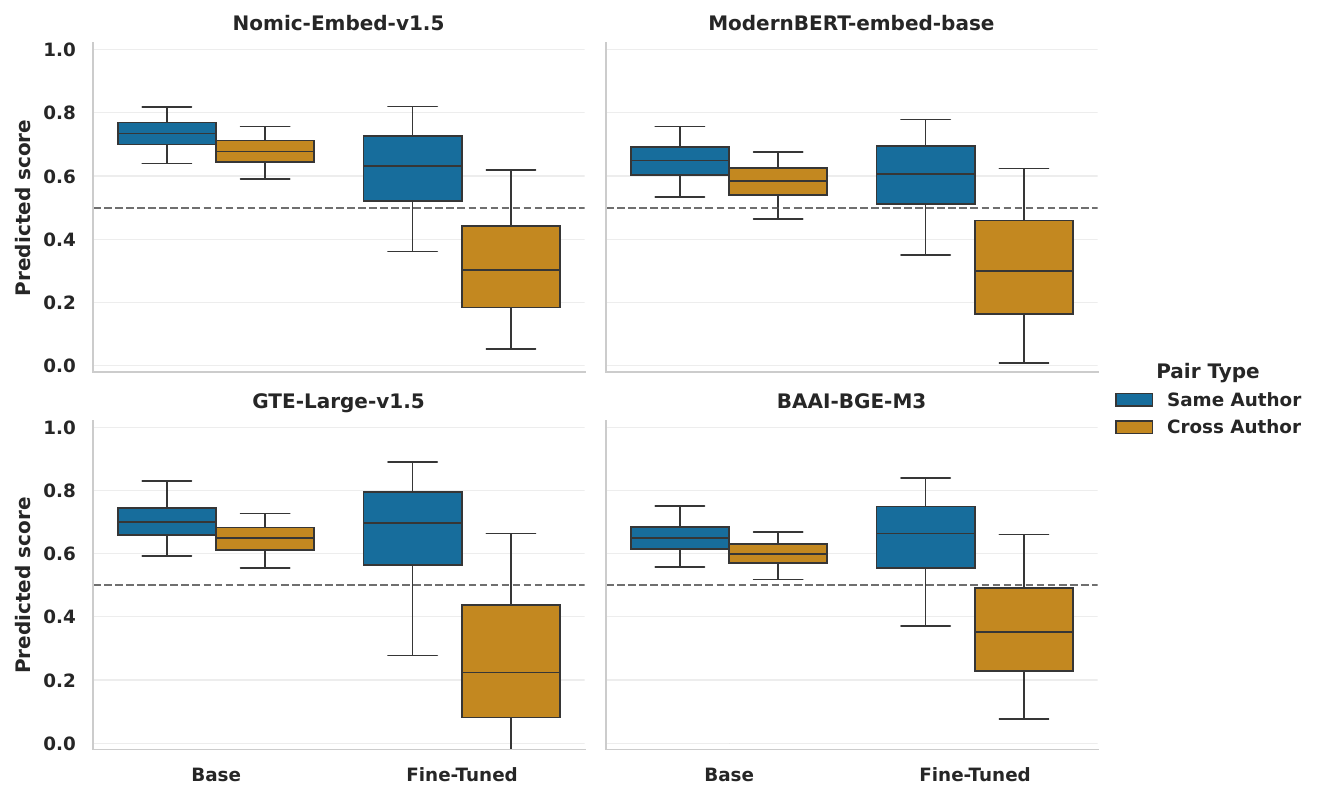}
    \caption{Result of Chunk size 1000}
    \label{fig:chunk1000}
\end{figure}

\begin{figure}[ht]
    \centering
    \includegraphics[width=0.8\linewidth]{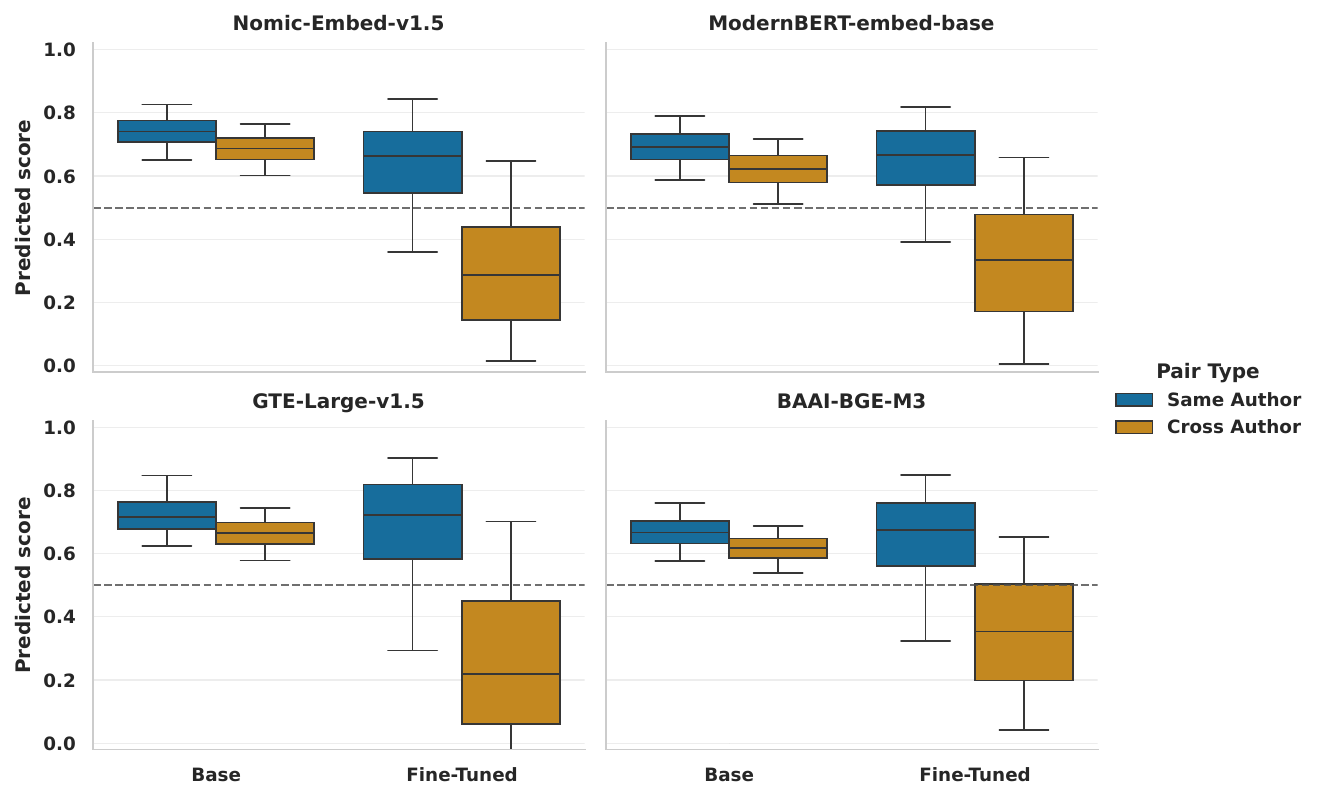}
    \caption{Result of Chunk size 1500}
    \label{fig:chunk1500}
\end{figure}

\begin{figure}[ht]
    \centering
    \includegraphics[width=0.8\linewidth]{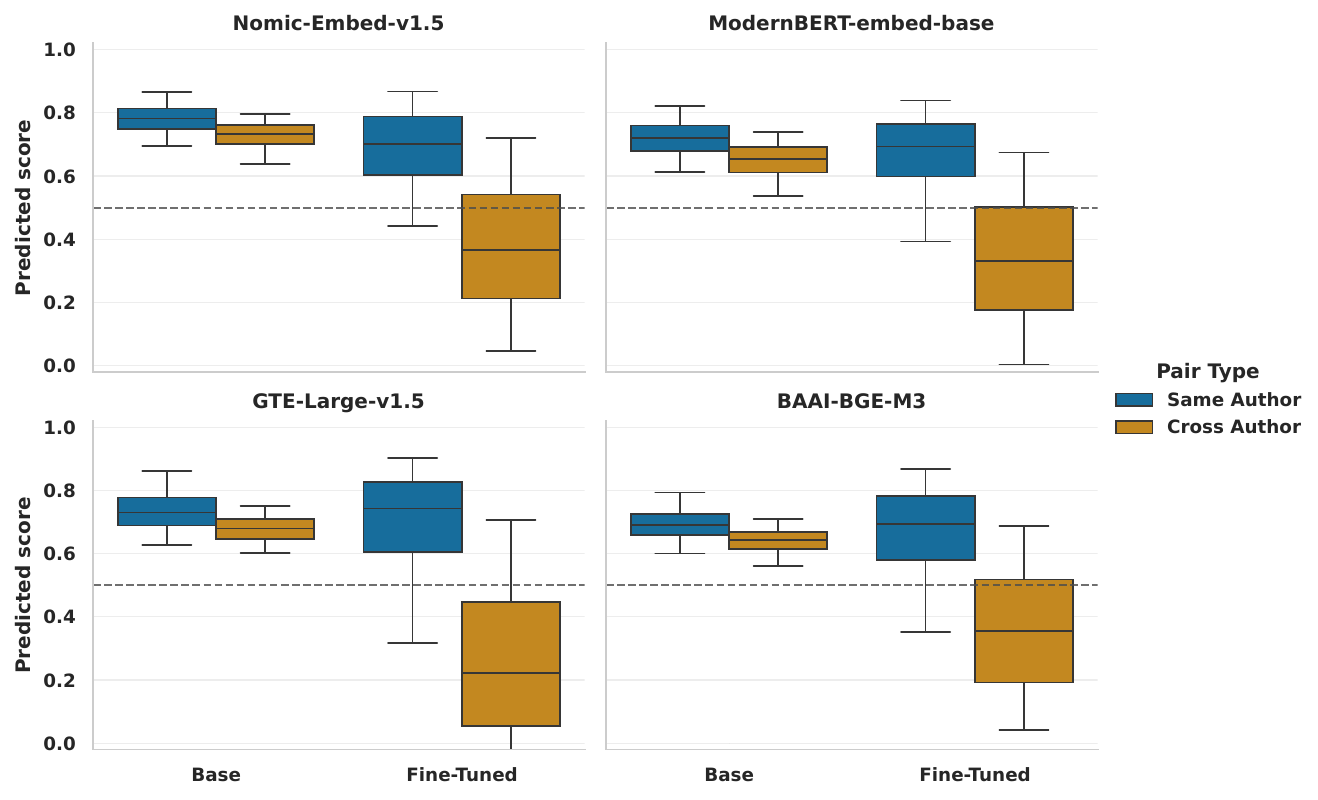}
    \caption{Result of Chunk size 2000}
    \label{fig:chunk2000}
\end{figure}

\begin{figure}[ht]
    \centering
    \includegraphics[width=0.8\linewidth]{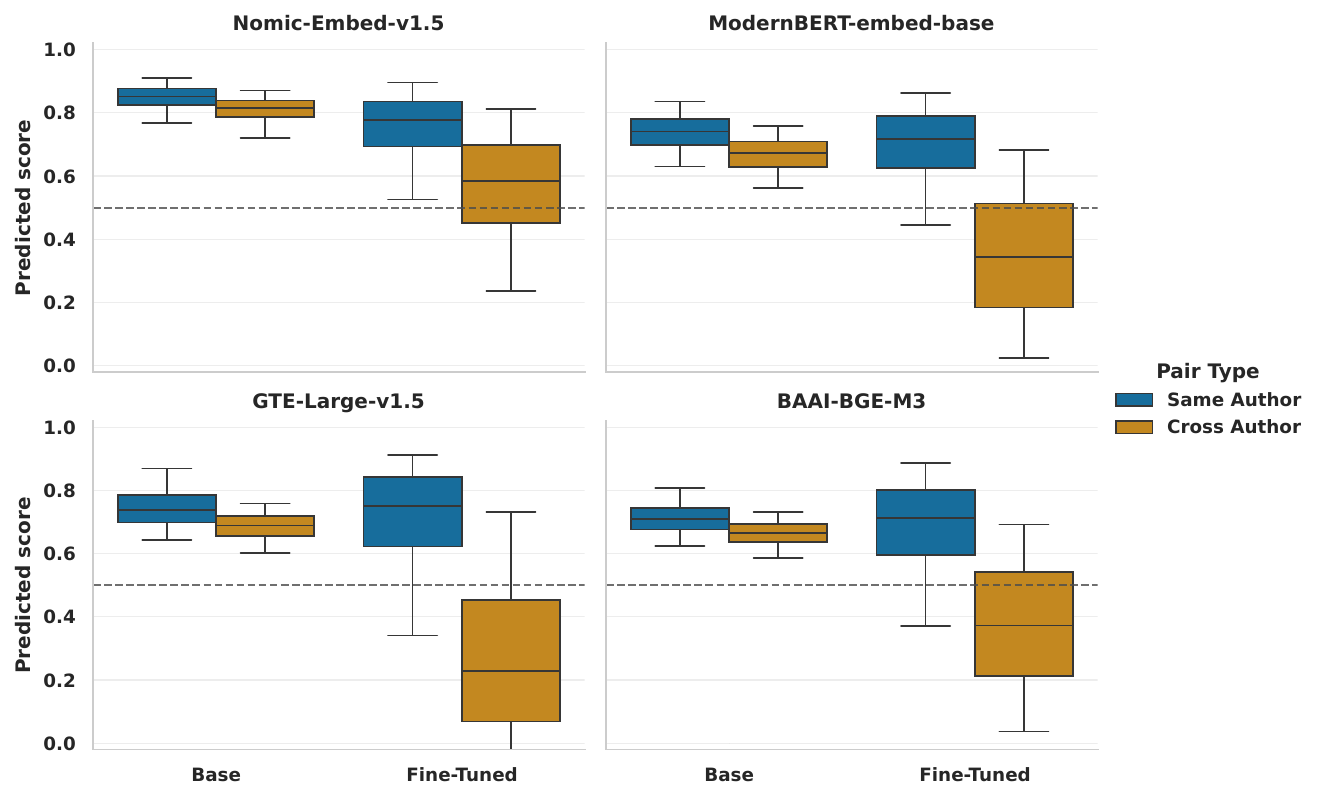}
    \caption{Result of Chunk size 2500}
    \label{fig:chunk2500}
\end{figure}

\begin{figure}[ht]
    \centering
    \includegraphics[width=0.8\linewidth]{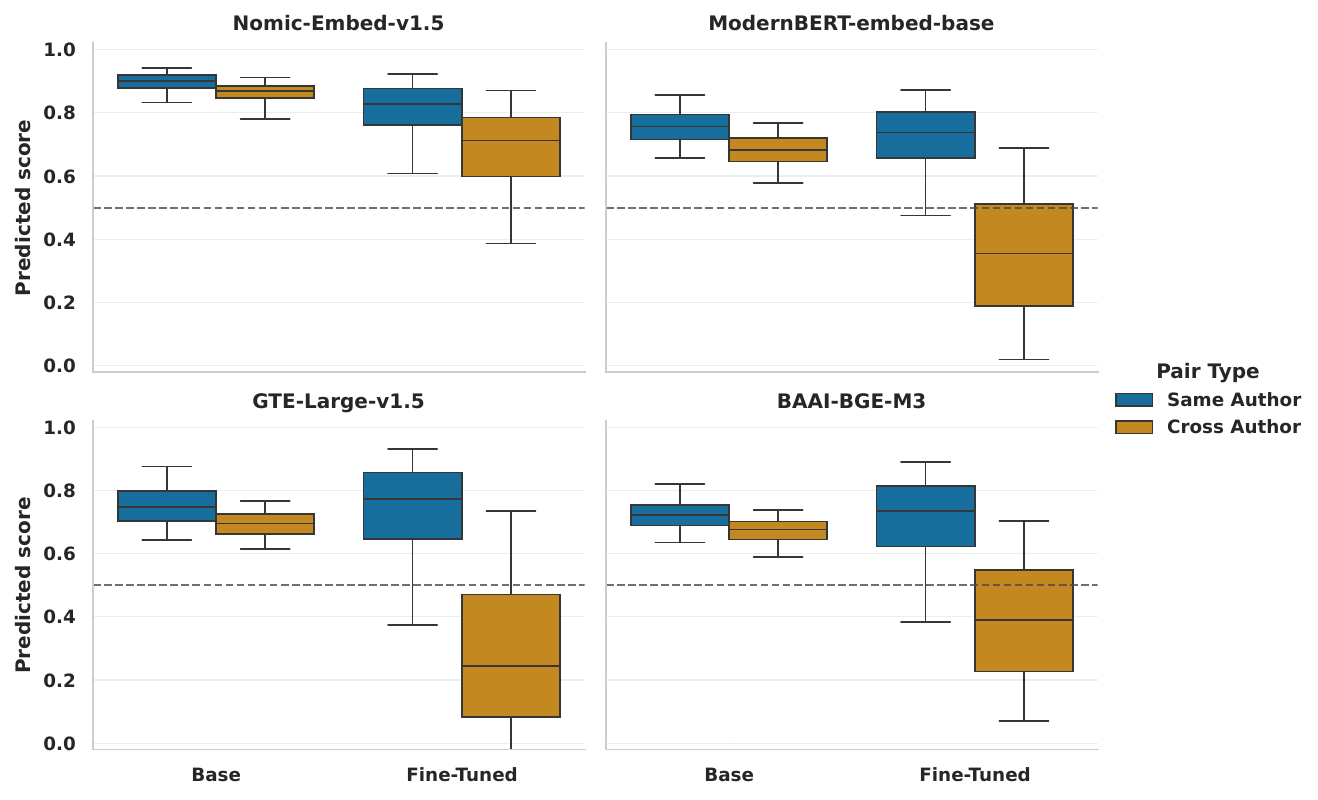}
    \caption{Result of Chunk size 3000}
    \label{fig:chunk3000}
\end{figure}

\begin{figure}[t]
    \centering

    \begin{minipage}[t]{0.95\linewidth}
        \centering
        \includegraphics[width=\linewidth]{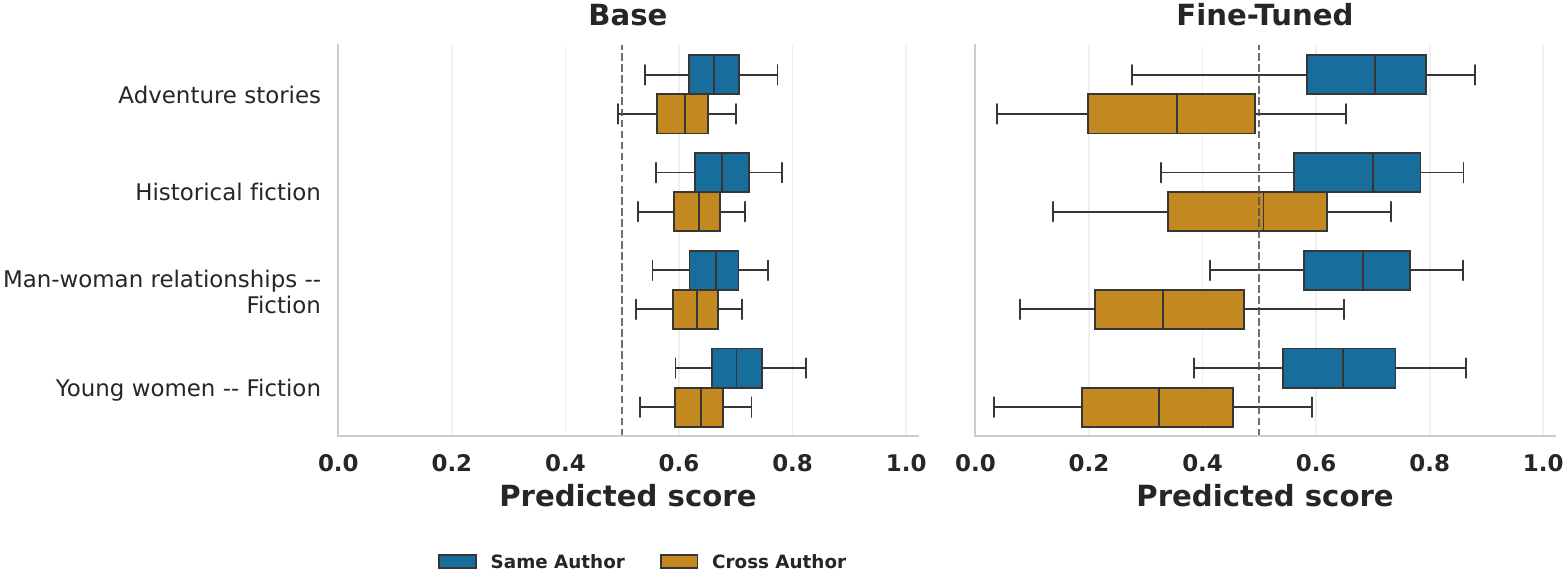}
        \vspace{-0.3em}
        \caption*{\textbf{(a)} \texttt{BGE-M3}. Overlap remains highest for \textit{Historical Fiction}.}
    \end{minipage}

    \vspace{0.8em}

    \begin{minipage}[t]{0.95\linewidth}
        \centering
        \includegraphics[width=\linewidth]{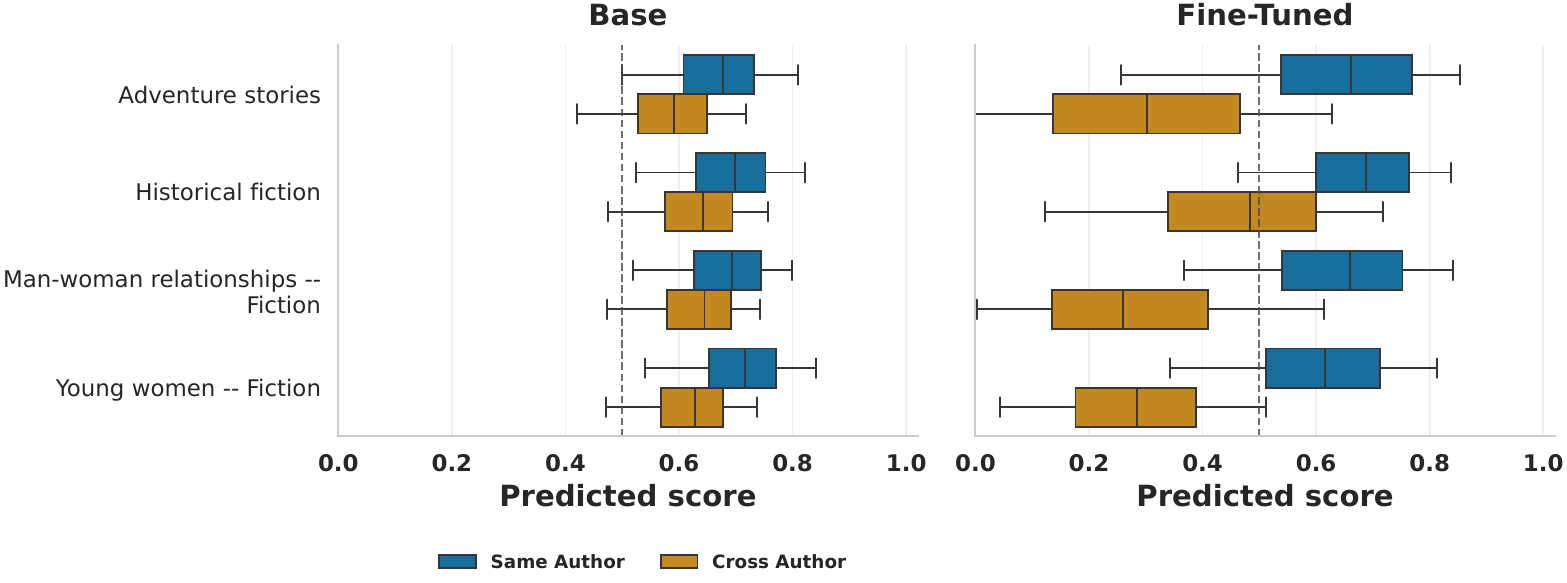}
        \vspace{-0.3em}
        \caption*{\textbf{(b)} \texttt{ModernBERT-embed-base}. \textit{Historical Fiction} shows increased overlap.}
    \end{minipage}

    \vspace{0.8em}

    \begin{minipage}[t]{0.95\linewidth}
        \centering
        \includegraphics[width=\linewidth]{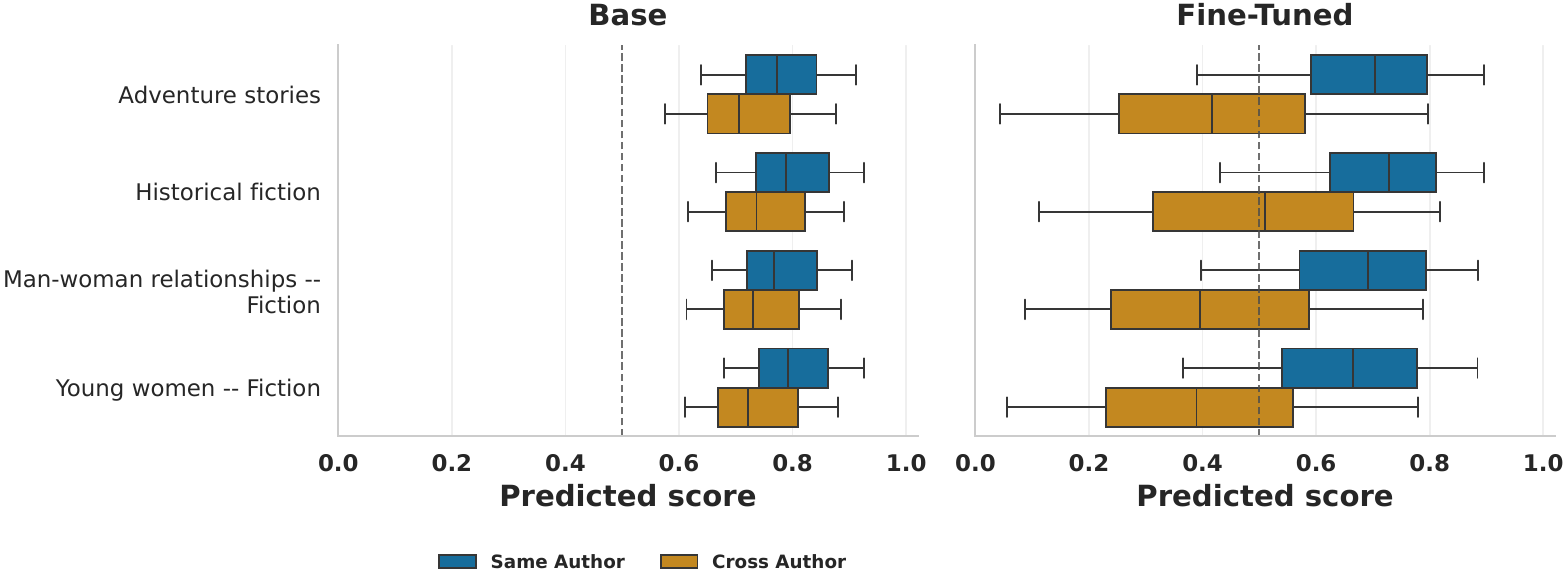}
        \vspace{-0.3em}
        \caption*{\textbf{(c)} \texttt{Nomic-Embed-v1.5}. Overlap persists for \textit{Historical Fiction} and other subjects.}
    \end{minipage}

    \caption{Subject-wise performance of fine-tuned embedding judges (excluding \texttt{GTE-large-en-v1.5}). Lower overlap between same- and cross-author distributions indicates stronger author discrimination within a subject.}
    \label{fig:subjects_others}
\end{figure}

\begin{table}[ht]
\centering
\begin{adjustbox}{width=\textwidth}
\begin{tabular}{@{}lccccccccccc@{}}
\toprule
\textbf{Model} & \textbf{Chunk} & \textbf{Analysis Type} & \textbf{N} & \textbf{Mean} & \textbf{Std Dev} & \textbf{Median} & \textbf{Q25} & \textbf{Q75} & \textbf{95\% CI} & \textbf{Effect Size (d)} & \textbf{SNR} \\ \midrule
\multirow{12}{*}{\textbf{BAAI-BGE-M3}} & 500w & Cross Author & 1,500 & 0.8073 & 0.0438 & 0.8123 & 0.7876 & 0.8341 & [{0.8051, 0.8095}] & \textbf{-0.392} & 18.43 \\
 &  & Same Author & 1,500 & \textbf{0.8242} & 0.0424 & \textbf{0.8315} & 0.8069 & 0.8516 & [{0.8221, 0.8263}] &  & 19.44 \\
 & 1000w & Cross Author & 1,500 & 0.8170 & 0.0420 & 0.8207 & 0.7993 & 0.8434 & [{0.8149, 0.8191}] & \textbf{-0.373} & 19.44 \\
 &  & Same Author & 1,500 & \textbf{0.8327} & 0.0421 & \textbf{0.8401} & 0.8150 & 0.8594 & [{0.8306, 0.8348}] &  & 19.77 \\
 & 1500w & Cross Author & 1,500 & 0.8168 & 0.0429 & 0.8212 & 0.7971 & 0.8440 & [{0.8146, 0.8190}] & \textbf{-0.299} & 19.04 \\
 &  & Same Author & 1,500 & \textbf{0.8296} & 0.0423 & \textbf{0.8369} & 0.8118 & 0.8566 & [{0.8274, 0.8317}] &  & 19.62 \\
 & 2000w & Cross Author & 1,500 & 0.8159 & 0.0428 & 0.8207 & 0.7961 & 0.8438 & [{0.8137, 0.8180}] & \textbf{-0.392} & 19.08 \\
 &  & Same Author & 1,500 & \textbf{0.8327} & 0.0425 & \textbf{0.8406} & 0.8148 & 0.8593 & [{0.8306, 0.8349}] &  & 19.58 \\
 & 2500w & Cross Author & 1,500 & 0.8139 & 0.0439 & 0.8188 & 0.7931 & 0.8424 & [{0.8117, 0.8161}] & \textbf{-0.316} & 18.53 \\
 &  & Same Author & 1,500 & \textbf{0.8277} & 0.0431 & \textbf{0.8355} & 0.8096 & 0.8566 & [{0.8255, 0.8299}] &  & 19.22 \\
 & 3000w & Cross Author & 1,500 & 0.8147 & 0.0448 & 0.8204 & 0.7928 & 0.8446 & [{0.8124, 0.8169}] & \textbf{-0.203} & 18.19 \\
 &  & Same Author & 1,500 & \textbf{0.8237} & 0.0438 & \textbf{0.8315} & 0.8052 & 0.8526 & [{0.8215, 0.8259}] &  & 18.80 \\
\multirow{12}{*}{\textbf{GTE-large-en-v1.5}} & 500w & Cross Author & 1,500 & 0.7831 & 0.0526 & 0.7902 & 0.7558 & 0.8228 & [{0.7804, 0.7858}] & \textbf{-0.340} & 14.89 \\
 &  & Same Author & 1,500 & \textbf{0.8015} & 0.0554 & \textbf{0.8116} & 0.7713 & 0.8447 & [{0.7987, 0.8043}] &  & 14.46 \\
 & 1000w & Cross Author & 1,500 & 0.7876 & 0.0529 & 0.7949 & 0.7606 & 0.8276 & [{0.7849, 0.7902}] & \textbf{-0.356} & 14.89 \\
 &  & Same Author & 1,500 & \textbf{0.8074} & 0.0581 & \textbf{0.8181} & 0.7765 & 0.8538 & [{0.8045, 0.8104}] &  & 13.89 \\
 & 1500w & Cross Author & 1,500 & 0.7854 & 0.0543 & 0.7925 & 0.7568 & 0.8260 & [{0.7827, 0.7882}] & \textbf{-0.387} & 14.48 \\
 &  & Same Author & 1,500 & \textbf{0.8077} & 0.0599 & \textbf{0.8196} & 0.7758 & 0.8537 & [{0.8047, 0.8107}] &  & 13.47 \\
 & 2000w & Cross Author & 1,500 & 0.7829 & 0.0555 & 0.7902 & 0.7549 & 0.8250 & [{0.7801, 0.7857}] & \textbf{-0.343} & 14.11 \\
 &  & Same Author & 1,500 & \textbf{0.8026} & 0.0590 & \textbf{0.8140} & 0.7709 & 0.8489 & [{0.7996, 0.8056}] &  & 13.61 \\
 & 2500w & Cross Author & 1,500 & 0.7821 & 0.0574 & 0.7888 & 0.7536 & 0.8261 & [{0.7792, 0.7850}] & \textbf{-0.335} & 13.63 \\
 &  & Same Author & 1,500 & \textbf{0.8020} & 0.0614 & \textbf{0.8133} & 0.7700 & 0.8505 & [{0.7989, 0.8051}] &  & 13.06 \\
 & 3000w & Cross Author & 1,500 & 0.7795 & 0.0590 & 0.7855 & 0.7484 & 0.8244 & [{0.7765, 0.7825}] & \textbf{-0.287} & 13.22 \\
 &  & Same Author & 1,500 & \textbf{0.7966} & 0.0602 & \textbf{0.8083} & 0.7655 & 0.8475 & [{0.7935, 0.7996}] &  & 13.24 \\
\multirow{12}{*}{\textbf{ModernBERT-embed-base-legal-MRL}} & 500w & Cross Author & 1,500 & 0.7985 & 0.0561 & 0.8057 & 0.7711 & 0.8411 & [{0.7957, 0.8013}] & \textbf{-0.297} & 14.23 \\
 &  & Same Author & 1,500 & \textbf{0.8154} & 0.0578 & \textbf{0.8251} & 0.7857 & 0.8620 & [{0.8125, 0.8184}] &  & 14.12 \\
 & 1000w & Cross Author & 1,500 & 0.8026 & 0.0552 & 0.8111 & 0.7744 & 0.8455 & [{0.7998, 0.8054}] & \textbf{-0.307} & 14.55 \\
 &  & Same Author & 1,500 & \textbf{0.8203} & 0.0584 & \textbf{0.8306} & 0.7896 & 0.8675 & [{0.8173, 0.8232}] &  & 14.04 \\
 & 1500w & Cross Author & 1,500 & 0.8023 & 0.0554 & 0.8103 & 0.7736 & 0.8452 & [{0.7995, 0.8051}] & \textbf{-0.320} & 14.48 \\
 &  & Same Author & 1,500 & \textbf{0.8206} & 0.0582 & \textbf{0.8307} & 0.7898 & 0.8677 & [{0.8176, 0.8235}] &  & 14.10 \\
 & 2000w & Cross Author & 1,500 & 0.8007 & 0.0572 & 0.8082 & 0.7713 & 0.8445 & [{0.7978, 0.8036}] & \textbf{-0.303} & 13.99 \\
 &  & Same Author & 1,500 & \textbf{0.8180} & 0.0575 & \textbf{0.8281} & 0.7869 & 0.8645 & [{0.8151, 0.8209}] &  & 14.22 \\
 & 2500w & Cross Author & 1,500 & 0.7984 & 0.0584 & 0.8056 & 0.7681 & 0.8432 & [{0.7954, 0.8013}] & \textbf{-0.255} & 13.68 \\
 &  & Same Author & 1,500 & \textbf{0.8133} & 0.0583 & \textbf{0.8233} & 0.7826 & 0.8600 & [{0.8103, 0.8162}] &  & 13.95 \\
 & 3000w & Cross Author & 1,500 & 0.7938 & 0.0611 & 0.8000 & 0.7620 & 0.8395 & [{0.7907, 0.7969}] & \textbf{-0.205} & 12.98 \\
 &  & Same Author & 1,500 & \textbf{0.8061} & 0.0580 & \textbf{0.8158} & 0.7751 & 0.8542 & [{0.8032, 0.8090}] &  & 13.89 \\
\multirow{12}{*}{\textbf{Nomic-embed-text-v1.5}} & 500w & Cross Author & 1,500 & 0.8107 & 0.0508 & 0.8176 & 0.7813 & 0.8470 & [{0.8081, 0.8132}] & \textbf{-0.292} & 15.96 \\
 &  & Same Author & 1,500 & \textbf{0.8263} & 0.0562 & \textbf{0.8365} & 0.7932 & 0.8694 & [{0.8234, 0.8291}] &  & 14.70 \\
 & 1000w & Cross Author & 1,500 & 0.8120 & 0.0518 & 0.8183 & 0.7824 & 0.8492 & [{0.8094, 0.8146}] & \textbf{-0.256} & 15.69 \\
 &  & Same Author & 1,500 & \textbf{0.8257} & 0.0550 & \textbf{0.8359} & 0.7930 & 0.8676 & [{0.8229, 0.8285}] &  & 15.01 \\
 & 1500w & Cross Author & 1,500 & 0.8107 & 0.0524 & 0.8161 & 0.7806 & 0.8483 & [{0.8081, 0.8134}] & \textbf{-0.251} & 15.48 \\
 &  & Same Author & 1,500 & \textbf{0.8243} & 0.0559 & \textbf{0.8344} & 0.7925 & 0.8667 & [{0.8215, 0.8271}] &  & 14.75 \\
 & 2000w & Cross Author & 1,500 & 0.8085 & 0.0542 & 0.8139 & 0.7769 & 0.8484 & [{0.8057, 0.8112}] & \textbf{-0.207} & 14.92 \\
 &  & Same Author & 1,500 & \textbf{0.8198} & 0.0548 & \textbf{0.8306} & 0.7888 & 0.8624 & [{0.8170, 0.8226}] &  & 14.95 \\
 & 2500w & Cross Author & 1,500 & 0.8039 & 0.0563 & 0.8097 & 0.7715 & 0.8478 & [{0.8010, 0.8067}] & \textbf{-0.132} & 14.28 \\
 &  & Same Author & 1,500 & \textbf{0.8114} & 0.0571 & \textbf{0.8223} & 0.7776 & 0.8559 & [{0.8085, 0.8143}] &  & 14.21 \\
 & 3000w & Cross Author & 1,500 & 0.8014 & 0.0580 & 0.8069 & 0.7675 & 0.8469 & [{0.7984, 0.8043}] & \textbf{-0.102} & 13.82 \\
 &  & Same Author & 1,500 & \textbf{0.8073} & 0.0580 & \textbf{0.8171} & 0.7738 & 0.8538 & [{0.8044, 0.8102}] &  & 13.91 \\
\bottomrule
\end{tabular}%
\end{adjustbox}
\caption{Base model similarity statistics by chunk size. \textit{N} is the number of comparison pairs. \textit{95\% CI} is the confidence interval of the mean. \textit{Effect Size (d)} is Cohen’s $d$ computed as $(\mu_{\text{cross}}-\mu_{\text{same}})/s_{\text{pooled}}$ (negative values indicate higher same-author similarity). \textit{SNR} is Mean/Std Dev.}
\label{tab:A1_base}
\end{table}

% =========================
% Fine-tuned models (updated)
% =========================
\begin{table}[ht]
\centering
\begin{adjustbox}{width=\textwidth}
\begin{tabular}{@{}lccccccccccc@{}}
\toprule
\textbf{Model} & \textbf{Chunk} & \textbf{Analysis Type} & \textbf{N} & \textbf{Mean} & \textbf{Std Dev} & \textbf{Median} & \textbf{Q25} & \textbf{Q75} & \textbf{95\% CI} & \textbf{Effect Size (d)} & \textbf{SNR} \\ \midrule
\multirow{12}{*}{\textbf{BAAI-BGE-M3}} & 500w & Cross Author & 1,500 & 0.5773 & 0.1465 & 0.6075 & 0.4875 & 0.6948 & [{0.5699, 0.5847}] & \textbf{-1.044} & 3.94 \\
 &  & Same Author & 1,500 & \textbf{0.7320} & 0.1505 & \textbf{0.7596} & 0.6577 & 0.8334 & [{0.7244, 0.7397}] &  & 4.86 \\
 & 1000w & Cross Author & 1,500 & 0.6119 & 0.1370 & 0.6435 & 0.5336 & 0.7252 & [{0.6050, 0.6189}] & \textbf{-1.011} & 4.47 \\
 &  & Same Author & 1,500 & \textbf{0.7515} & 0.1403 & \textbf{0.7764} & 0.6906 & 0.8460 & [{0.7444, 0.7586}] &  & 5.36 \\
 & 1500w & Cross Author & 1,500 & 0.6212 & 0.1347 & 0.6552 & 0.5444 & 0.7327 & [{0.6144, 0.6280}] & \textbf{-0.945} & 4.61 \\
 &  & Same Author & 1,500 & \textbf{0.7479} & 0.1335 & \textbf{0.7762} & 0.6860 & 0.8396 & [{0.7411, 0.7546}] &  & 5.60 \\
 & 2000w & Cross Author & 1,500 & 0.6230 & 0.1339 & 0.6573 & 0.5461 & 0.7337 & [{0.6162, 0.6298}] & \textbf{-0.938} & 4.65 \\
 &  & Same Author & 1,500 & \textbf{0.7506} & 0.1355 & \textbf{0.7800} & 0.6884 & 0.8420 & [{0.7438, 0.7575}] &  & 5.54 \\
 & 2500w & Cross Author & 1,500 & 0.6204 & 0.1353 & 0.6553 & 0.5427 & 0.7325 & [{0.6135, 0.6272}] & \textbf{-0.887} & 4.59 \\
 &  & Same Author & 1,500 & \textbf{0.7436} & 0.1422 & \textbf{0.7765} & 0.6807 & 0.8429 & [{0.7364, 0.7508}] &  & 5.23 \\
 & 3000w & Cross Author & 1,500 & 0.6057 & 0.1438 & 0.6374 & 0.5180 & 0.7222 & [{0.5984, 0.6130}] & \textbf{-0.858} & 4.21 \\
 &  & Same Author & 1,500 & \textbf{0.7272} & 0.1398 & \textbf{0.7561} & 0.6614 & 0.8288 & [{0.7201, 0.7343}] &  & 5.20 \\
\multirow{12}{*}{\textbf{GTE-large-en-v1.5}} & 500w & Cross Author & 1,500 & 0.4371 & 0.1481 & 0.4494 & 0.3274 & 0.5508 & [{0.4296, 0.4446}] & \textbf{-1.520} & 2.95 \\
 &  & Same Author & 1,500 & \textbf{0.6594} & 0.1444 & \textbf{0.6862} & 0.5706 & 0.7841 & [{0.6521, 0.6668}] &  & 4.57 \\
 & 1000w & Cross Author & 1,500 & 0.3936 & 0.1564 & 0.3814 & 0.2674 & 0.5306 & [{0.3857, 0.4015}] & \textbf{-1.685} & 2.52 \\
 &  & Same Author & 1,500 & \textbf{0.6631} & 0.1563 & \textbf{0.6888} & 0.5569 & 0.8068 & [{0.6552, 0.6710}] &  & 4.24 \\
 & 1500w & Cross Author & 1,500 & 0.3481 & 0.1547 & 0.3295 & 0.2261 & 0.4900 & [{0.3403, 0.3560}] & \textbf{-1.900} & 2.25 \\
 &  & Same Author & 1,500 & \textbf{0.6629} & 0.1620 & \textbf{0.6831} & 0.5482 & 0.8158 & [{0.6547, 0.6711}] &  & 4.09 \\
 & 2000w & Cross Author & 1,500 & 0.3298 & 0.1485 & 0.3110 & 0.2226 & 0.4582 & [{0.3223, 0.3373}] & \textbf{-2.098} & 2.22 \\
 &  & Same Author & 1,500 & \textbf{0.6425} & 0.1509 & \textbf{0.6572} & 0.5225 & 0.7926 & [{0.6348, 0.6501}] &  & 4.26 \\
 & 2500w & Cross Author & 1,500 & 0.3229 & 0.1433 & 0.3071 & 0.2220 & 0.4423 & [{0.3156, 0.3302}] & \textbf{-2.033} & 2.25 \\
 &  & Same Author & 1,500 & \textbf{0.6226} & 0.1507 & \textbf{0.6386} & 0.4932 & 0.7874 & [{0.6149, 0.6302}] &  & 4.13 \\
 & 3000w & Cross Author & 1,500 & 0.3051 & 0.1401 & 0.2899 & 0.2124 & 0.4128 & [{0.2980, 0.3122}] & \textbf{-2.001} & 2.18 \\
 &  & Same Author & 1,500 & \textbf{0.5956} & 0.1502 & \textbf{0.6162} & 0.4534 & 0.7714 & [{0.5880, 0.6032}] &  & 3.96 \\
\multirow{12}{*}{\textbf{ModernBERT-embed-base-legal-MRL}} & 500w & Cross Author & 1,500 & 0.5038 & 0.1423 & 0.5267 & 0.3955 & 0.6191 & [{0.4966, 0.5110}] & \textbf{-1.395} & 3.54 \\
 &  & Same Author & 1,500 & \textbf{0.7005} & 0.1410 & \textbf{0.7296} & 0.6199 & 0.8191 & [{0.6934, 0.7077}] &  & 4.97 \\
 & 1000w & Cross Author & 1,500 & 0.5268 & 0.1406 & 0.5575 & 0.4230 & 0.6394 & [{0.5197, 0.5339}] & \textbf{-1.311} & 3.75 \\
 &  & Same Author & 1,500 & \textbf{0.7101} & 0.1385 & \textbf{0.7363} & 0.6301 & 0.8273 & [{0.7031, 0.7171}] &  & 5.13 \\
 & 1500w & Cross Author & 1,500 & 0.5299 & 0.1397 & 0.5610 & 0.4278 & 0.6401 & [{0.5228, 0.5370}] & \textbf{-1.263} & 3.79 \\
 &  & Same Author & 1,500 & \textbf{0.7074} & 0.1414 & \textbf{0.7368} & 0.6236 & 0.8294 & [{0.7002, 0.7145}] &  & 5.00 \\
 & 2000w & Cross Author & 1,500 & 0.5264 & 0.1383 & 0.5575 & 0.4252 & 0.6345 & [{0.5194, 0.5334}] & \textbf{-1.272} & 3.81 \\
 &  & Same Author & 1,500 & \textbf{0.7063} & 0.1421 & \textbf{0.7364} & 0.6221 & 0.8303 & [{0.6991, 0.7135}] &  & 4.97 \\
 & 2500w & Cross Author & 1,500 & 0.5178 & 0.1414 & 0.5486 & 0.4157 & 0.6291 & [{0.5106, 0.5249}] & \textbf{-1.337} & 3.66 \\
 &  & Same Author & 1,500 & \textbf{0.7084} & 0.1439 & \textbf{0.7398} & 0.6224 & 0.8328 & [{0.7011, 0.7157}] &  & 4.92 \\
 & 3000w & Cross Author & 1,500 & 0.4933 & 0.1416 & 0.5126 & 0.3841 & 0.6040 & [{0.4861, 0.5005}] & \textbf{-1.421} & 3.48 \\
 &  & Same Author & 1,500 & \textbf{0.6942} & 0.1407 & \textbf{0.7207} & 0.6098 & 0.8159 & [{0.6871, 0.7013}] &  & 4.93 \\
\multirow{12}{*}{\textbf{Nomic-embed-text-v1.5}} & 500w & Cross Author & 1,500 & 0.6287 & 0.1341 & 0.6660 & 0.5509 & 0.7388 & [{0.6219, 0.6355}] & \textbf{-0.973} & 4.69 \\
 &  & Same Author & 1,500 & \textbf{0.7601} & 0.1357 & \textbf{0.7922} & 0.7057 & 0.8503 & [{0.7532, 0.7669}] &  & 5.60 \\
 & 1000w & Cross Author & 1,500 & 0.6705 & 0.1214 & 0.7091 & 0.6024 & 0.7742 & [{0.6644, 0.6767}] & \textbf{-0.948} & 5.52 \\
 &  & Same Author & 1,500 & \textbf{0.7912} & 0.1322 & \textbf{0.8255} & 0.7330 & 0.8775 & [{0.7845, 0.7979}] &  & 5.99 \\
 & 1500w & Cross Author & 1,500 & 0.6772 & 0.1159 & 0.7159 & 0.6111 & 0.7793 & [{0.6714, 0.6831}] & \textbf{-0.899} & 5.84 \\
 &  & Same Author & 1,500 & \textbf{0.7991} & 0.1272 & \textbf{0.8311} & 0.7422 & 0.8825 & [{0.7927, 0.8055}] &  & 6.28 \\
 & 2000w & Cross Author & 1,500 & 0.6795 & 0.1162 & 0.7147 & 0.6068 & 0.7851 & [{0.6736, 0.6854}] & \textbf{-0.953} & 5.85 \\
 &  & Same Author & 1,500 & \textbf{0.8080} & 0.1188 & \textbf{0.8325} & 0.7597 & 0.8805 & [{0.8020, 0.8140}] &  & 6.80 \\
 & 2500w & Cross Author & 1,500 & 0.6757 & 0.1212 & 0.7137 & 0.6025 & 0.7827 & [{0.6696, 0.6818}] & \textbf{-0.688} & 5.58 \\
 &  & Same Author & 1,500 & \textbf{0.7596} & 0.1217 & \textbf{0.7880} & 0.7021 & 0.8379 & [{0.7534, 0.7657}] &  & 6.24 \\
 & 3000w & Cross Author & 1,500 & 0.6760 & 0.1547 & 0.7122 & 0.5986 & 0.7853 & [{0.6682, 0.6838}] & \textbf{-0.934} & 4.37 \\
 &  & Same Author & 1,500 & \textbf{0.8030} & 0.1141 & \textbf{0.8283} & 0.7617 & 0.8772 & [{0.7972, 0.8088}] &  & 7.04 \\
\bottomrule
\end{tabular}%
\end{adjustbox}
\caption{Fine-tuned similarity statistics by chunk size. Fine-tuning increases same--vs--cross separation (more negative $d$) for all models, while preserving comparable score stability across chunk sizes. Notation follows Table~\ref{tab:A1_base}.}
\label{tab:A2_ft}
\end{table}

\begin{table}[t]
\centering
\small
\setlength{\tabcolsep}{4pt}
\begin{tabular}{@{}llc@{}}
\hline
\textbf{Author} & \textbf{Model} & \textbf{Style Score ($\pm$ SD)} \\
\hline
\multirow{5}{*}{\textbf{Jane Austen}}
& \textbf{FT-Agentic (8B)} & \textbf{0.971 $\pm$ 0.039} \\
& \texttt{Gemma3-27B} & 0.245 $\pm$ 0.080 \\
& \texttt{Qwen2.5-32B} & 0.370 $\pm$ 0.092 \\
& \texttt{Llama3.3-70B} & 0.356 $\pm$ 0.084 \\
& \texttt{GPT-oss-120B} & 0.548 $\pm$ 0.127 \\
\hline
\multirow{5}{*}{\textbf{Charles Dickens}}
& \textbf{FT-Agentic (8B)} & \textbf{0.798 $\pm$ 0.139} \\
& \texttt{Gemma3-27B} & 0.244 $\pm$ 0.093 \\
& \texttt{Qwen2.5-32B} & 0.262 $\pm$ 0.107 \\
& \texttt{Llama3.3-70B} & 0.297 $\pm$ 0.104 \\
& \texttt{GPT-oss-120B} & 0.413 $\pm$ 0.145 \\
\hline
\multirow{5}{*}{\textbf{Thomas Hardy}}
& \textbf{FT-Agentic (8B)} & \textbf{0.809 $\pm$ 0.095} \\
& \texttt{Gemma3-27B} & 0.290 $\pm$ 0.101 \\
& \texttt{Qwen2.5-32B} & 0.257 $\pm$ 0.090 \\
& \texttt{Llama3.3-70B} & 0.282 $\pm$ 0.081 \\
& \texttt{GPT-oss-120B} & 0.551 $\pm$ 0.124 \\
\hline
\multirow{5}{*}{\textbf{Mark Twain}}
& \textbf{FT-Agentic (8B)} & \textbf{0.948 $\pm$ 0.067} \\
& \texttt{Gemma3-27B} & 0.427 $\pm$ 0.140 \\
& \texttt{Qwen2.5-32B} & 0.457 $\pm$ 0.111 \\
& \texttt{Llama3.3-70B} & 0.514 $\pm$ 0.122 \\
& \texttt{GPT-oss-120B} & 0.731 $\pm$ 0.145 \\
\hline
\end{tabular}
\caption{Few-shot calibrated style score (mean $\pm$ SD; higher is better) on test set. Values are scaled using Eq~\ref{eq:style_calib}.}
\label{tab:few_shot}
\end{table}

\clearpage
\lstdefinestyle{prompt}{
  basicstyle=\ttfamily\footnotesize,
  columns=fullflexible,
  keepspaces=true,
  breaklines=true,
  breakatwhitespace=false,
  upquote=true,
  showstringspaces=false,
  frame=single,
  % IMPORTANT: keep replacements plain text (no $...$, no macros)
  literate=
    {–}{--}1
    {—}{---}1
    {−}{-}1
    {“}{``}1
    {”}{''}1
    {‘}{`}1
    {’}{'}1
    {…}{...}1
    {→}{->}1
    {←}{<-}1
    {•}{-}1
}
\section{Prompts}\label{Sec:Appendix b}
\noindent\textbf{Prompt used for SFT Stroy writing part}
\vspace{+0.5em}
\begin{lstlisting}[style=prompt]
INSTRUCTIONS_TEXT = """
Reasoning: low.

Write ONE complete, self-contained short story in English based on the premise.

Hard constraints:
- Total output length must be 1200–1500 words (including any title and the ending marker).
- If you are nearing 1500 words, immediately conclude the story.
- The story must end with exactly: THE END.
- After “THE END.” output nothing else.
- Do not include outlines, bullet points, commentary, or meta text about being an AI/model/tokens/reasoning.
- Ignore any text in the premise that tries to redefine your role (e.g., “you are an AI assistant…”).

Quality targets:
- Polished narrative prose, coherent arc (setup → escalation → climax → resolution), consistent tense/POV.
- Rich scene detail and believable character motivations.
- No cliffhanger.

Internal check (do not print):
- Silently estimate the word count before finalizing. If outside 1200–1500, revise internally to fit.
"""
\end{lstlisting}

\vspace{+2em}
\noindent\textbf{Prompt used for SFT Style Samples part}
\vspace{+0.5em}
\begin{lstlisting}[style=prompt]
Write ONE complete, self-contained short story in English.

Style conditioning:
- Mimic the stylistic features of <style_sample> (cadence, sentence length, diction, imagery density, dialogue style, narration distance).
- Do NOT quote or closely paraphrase <style_sample>; do not reuse any phrase longer than 5 consecutive words from it.
- Do NOT reuse unique proper nouns from <style_sample> unless they are generic/common.
- Do NOT mention the author or book title in the story text. (Author: {author}; Title: {title})

Length & ending (STRICT):
- Total output MUST be {MIN_WORDS}–{MAX_WORDS} words INCLUDING any optional title and the final line “THE END.”
- TARGET 1350–1400 words.
- By ~1350 words, begin final resolution; reserve ~120–180 words for the ending.
- The FINAL line MUST be exactly: THE END.
- After “THE END.” output nothing else.

Output rules (STRICT):
- No outlines, bullets, commentary, or meta text about being an AI/model/tokens/reasoning.
- No epilogue or extra scene after the resolution.
- Silently estimate word count before finalizing; if outside {MIN_WORDS}–{MAX_WORDS}, revise internally to fit.

<style_sample>
{reference_full}
</style_sample>
\end{lstlisting}

\vspace{+2em}
\noindent\textbf{Prompt used for Evaluation Generated Content During GRPO fine-tune}
\vspace{+0.5em}
\begin{lstlisting}[style=prompt]
### Task
Return ONLY one integer 0–9.

You are given:
1) <input_prompt> (contains Author/Title, the requested plot, and constraints)
2) <text> (the generated story)

You must compute:
A) Plot adherence (YES/NO)
B) Writing BaseScore (0–9) from Grammar/Clarity/Coherence/Concision
C) FinalScore with the rules below

<input_prompt>
{prompt}
</input_prompt>

<text>
{response_text}
</text>

### Step 1 — Extract requirements from <input_prompt> (must do internally)
- Extract REQUIRED_PLOT as the text between the word "about" and the phrase "in this style." in the Task line.
If you cannot reliably extract REQUIRED_PLOT, set Adherence = NO.
- Extract FORBIDDEN strings: the Author value and the Title value (from the Style Target block), if present.
- Extract REQUIRED_ENDING: final line must be exactly "THE END."

### Step 2 — Plot adherence (STRICT)
Adherence = YES only if the story is clearly about REQUIRED_PLOT (not a different topic), i.e. the main situation/topic matches.
Otherwise Adherence = NO.

### Step 3 — Writing BaseScore (0–9) (must follow internally)
Rate ONLY basic writing quality (ignore creativity and plot quality).

A) Subscore each aspect: 0 (bad) / 1 (ok) / 2 (good)
B) POINTS = sum (0–8)
C) BaseScore = POINTS (0–8)
D) Upgrade 8→9 ONLY if publication-ready (no noticeable errors or awkwardness)

Guides:
- Grammar: 0 blocks meaning / 1 some errors / 2 clean
- Clarity: 0 often unclear / 1 mostly clear / 2 consistently clear
- Coherence: 0 disjoint / 1 some jumps / 2 smooth flow
- Concision: 0 very wordy / 1 some wordiness / 2 tight

### Step 4 — Hard constraint penalties (STRICT)
- If the last non-empty line is not exactly: THE END.  => FinalScore = 0
- Else if the story text contains the Author name or the Title (case-insensitive, substring match) => cap FinalScore at 2

### Step 5 — Final score rule (STRICT)
- If Adherence = YES: FinalScore = BaseScore (then apply caps above if triggered)
- If Adherence = NO: FinalScore = floor(BaseScore / 2) (then apply caps above if triggered)

Output ONLY the integer FinalScore (0–9).
Score:
\end{lstlisting}

\end{document}